\documentclass{article}
\usepackage{microtype}
\usepackage{graphicx}
\usepackage{subfigure}
\usepackage{booktabs} 

\usepackage{courier} 
\usepackage{pifont}
\usepackage{hyperref}
\usepackage{url}
\usepackage{mathtools}
\usepackage[utf8]{inputenc}
\usepackage[T1]{fontenc}
\usepackage{amsfonts}
\usepackage{nicefrac}
\usepackage{microtype}
\usepackage{xcolor}
\usepackage{multirow}
\usepackage{subcaption}
\usepackage{xspace}
\usepackage{wrapfig}
\usepackage{colortbl}
\usepackage[most]{tcolorbox}
\usepackage[ruled, vlined]{algorithm2e}
\usepackage{amsmath}
\usepackage{fontawesome}
\usepackage{tabularx}
\usepackage{amssymb}

\definecolor{citecolor}{HTML}{27ae60}
\definecolor{linkcolor}{HTML}{27ae60}
\hypersetup{breaklinks=true,colorlinks, linkcolor=linkcolor, citecolor=citecolor}

\usepackage{hyperref}

\usepackage[accepted]{icml2025}

\usepackage{amsthm}

\usepackage[capitalize,noabbrev]{cleveref}

\theoremstyle{plain}

\theoremstyle{definition}

\theoremstyle{remark}

\usepackage[textsize=tiny]{todonotes}

\usepackage{listings}

\usepackage{relsize}
\lstdefinestyle{textFileStyle}{
  basicstyle=\scriptsize\ttfamily,        
  breaklines=true,
backgroundcolor=\color{gray!5}, 
    frame=single, 
  numbers=none,                    
  breakindent=0pt,
  numbersep=5pt,                   
  numberstyle=\tiny\color{gray}, 
  rulecolor=\color{black},         
  showspaces=false,                
  showstringspaces=false,          
  showtabs=false,                  
  stringstyle=\color{purple},     
  tabsize=4,	                   
  title=\lstname,                   
  xleftmargin=.2cm,
  xrightmargin=.2cm,
  aboveskip=1.5em,
  belowskip=1.5 \baselineskip,
  belowcaptionskip=4pt
  }
\lstdefinestyle{python}{ 
  backgroundcolor=\color{white},   
  basicstyle=\scriptsize\ttfamily,        
  breakatwhitespace=false,         
  breaklines=true,                 
  captionpos=b,                    
  commentstyle=\color{green},    
  escapeinside={\%*}{*)},          
  extendedchars=false,              
  frame=lrtb,	                   
  keepspaces=true,                 
  keywordstyle=\color{blue},       
  language=Python,
  morekeywords={with,as},
  numbers=none,                    
  numbersep=5pt,                   
  numberstyle=\tiny\color{gray}, 
  rulecolor=\color{black},         
  showspaces=false,                
  showstringspaces=false,          
  showtabs=false,                  
  stringstyle=\color{purple},     
  tabsize=4,	                   
  title=\lstname,                   
  xleftmargin=.2cm,
  xrightmargin=.2cm,
  aboveskip=1.2em,
  belowskip=-1.5 \baselineskip,
  belowcaptionskip=0em,
}

\newcommand{\lmtt}[1]{\fontfamily{lmtt}\selectfont{#1}}

\newcommand{\ours}{{\lmtt NNetNav}\xspace}
\newcommand{\lourswa}{{\lmtt Llama8B-NNetNav-WA}\xspace}
\newcommand{\loursow}{{\lmtt Llama8B-NNetNav-Live}\xspace}
\newcommand{\loursb}{{\lmtt Llama8B-NNetNav-All}\xspace}

\newcommand{\eg}{\textit{e.g.}\xspace}

\newcommand{\hrow}{\rowcolor{gray!15}}

\newcommand\sms{\bgroup\markoverwith{\textcolor{blue}{\rule[.4ex]{2pt}{0.8pt}}}\ULon}

\newcommand{\exploref}{\ensuremath{\pi_\text{explore}}}
\newcommand{\labelf}{\ensuremath{\mathrm{Lf}_\text{LM}}}
\newcommand{\deltaf}{\ensuremath{\Delta_\text{LM}}}
\newcommand{\rewardf}{\ensuremath{s_\text{LM}}}

\newcommand{\policy}{\ensuremath{\pi_\text{LM}}}

\newcommand{\cmark}{\ding{51}}%
\newcommand{\xmark}{\ding{55}}%

\newcommand{\gpttl}{{\texttt{GPT-4o}\xspace}}

\newcommand{\llamas}{\texttt{Llama-3.1-8b}\xspace}
\newcommand{\llama}{\texttt{Llama-3.1-70b}\xspace}

\icmltitlerunning{NNetNav: Unsupervised Learning of Browser Agents Through Environment Interaction in the Wild}

\begin{document}
\twocolumn[
\icmltitle{
NNetNav: Unsupervised Learning of Browser Agents Through Environment Interaction in the Wild
}



\icmlsetsymbol{equal}{*}

\begin{icmlauthorlist}
\icmlauthor{Shikhar Murty}{yyy}
\icmlauthor{Hao Zhu}{yyy}
\icmlauthor{Dzmitry Bahdanau}{comp}
\icmlauthor{Christopher D. Manning}{yyy}

\end{icmlauthorlist}

\icmlaffiliation{yyy}{Computer Science Department, Stanford University}
\icmlaffiliation{comp}{ServiceNow Research}

\icmlcorrespondingauthor{Shikhar Murty}{smurty@cs.stanford.edu}

\icmlkeywords{Machine Learning, ICML}

\vskip 0.3in
]



\printAffiliationsAndNotice{}  

\begin{abstract}
We introduce \ours{}, a method for unsupervised interaction with websites that generates synthetic demonstrations for training browser agents. Given any website, \ours{} produces these demonstrations by retroactively labeling action sequences from an exploration policy. Most work on training browser agents has relied on expensive human supervision, and the limited prior work on such interaction-based techniques has failed to provide effective search through the exponentially large space of exploration. In contrast, \ours exploits the hierarchical structure of language instructions to make this search more tractable: Complex instructions are typically decomposable into simpler sub-tasks, allowing \ours to automatically prune interaction episodes when an intermediate trajectory cannot be annotated with a meaningful sub-task. \texttt{LLama-3.1-8b} finetuned on 10k \ours{} self-generated demonstrations obtains over 16\% success rate on WebArena, and 35\% on WebVoyager, an improvement of 15pts and 31pts respectively over zero-shot \texttt{LLama-3.1-8b}, outperforming zero-shot GPT-4 and reaching the state-of-the-art among unsupervised methods, for both benchmarks.

\end{abstract}

\section{Introduction}
Building grounded agents that map human language instructions to a sequence of executable actions is a long-standing goal of artificial intelligence \citep{winograd1972understand}. A promising new approach for building such agents is to use large language models to control policies in environments like web-browsers and computers \citep[among others]{yao2022react,  murty24bagel, xie2024osworld}.

\begin{figure}[!ht]
    \centering
\includegraphics[width=0.88\linewidth]{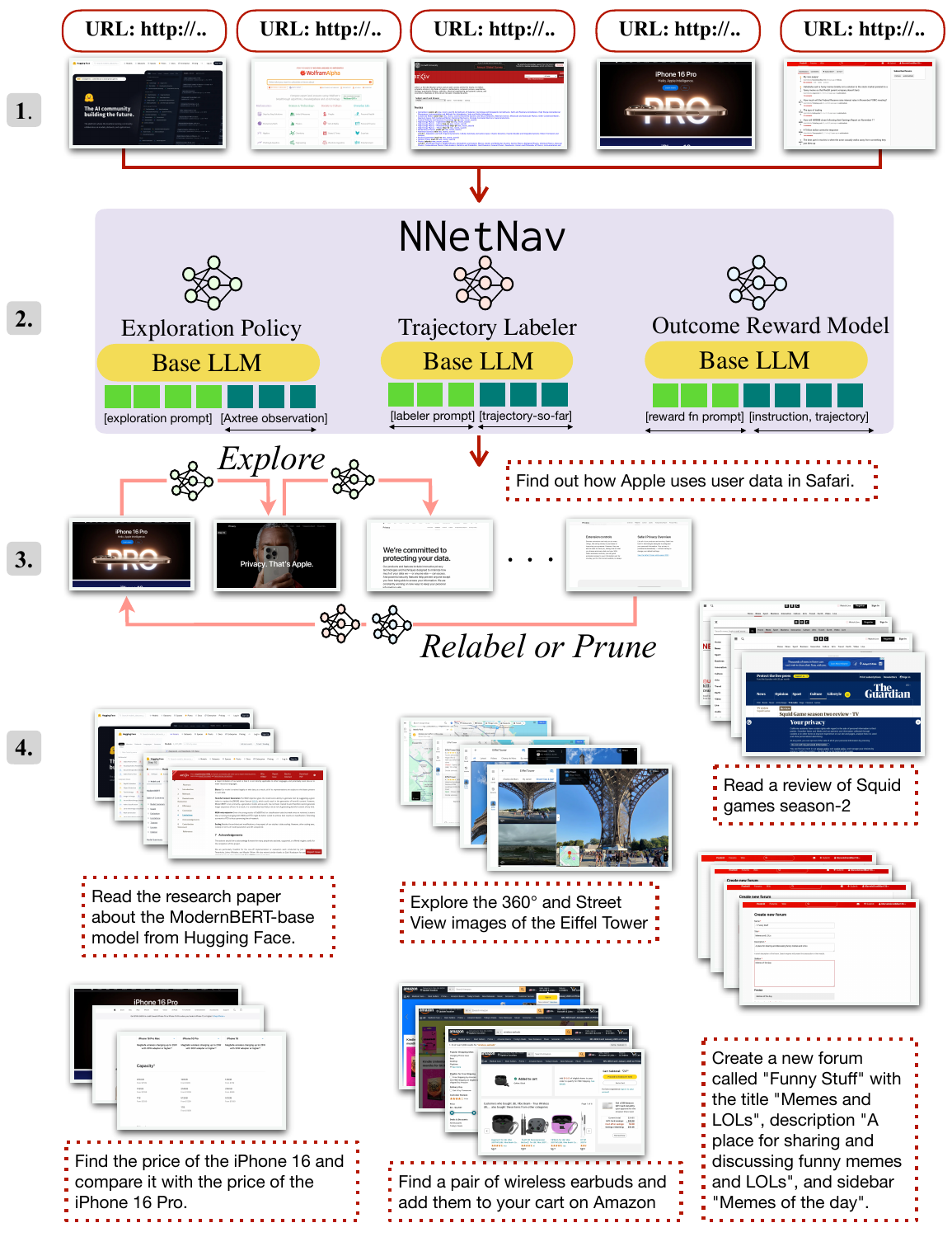}
\caption{Given web URLs (1), \ours{} (2) uses a structured exploration strategy to interact with websites (3) and autonomously discover diverse (instruction, trajectory) demonstrations, as summarized in (4).  To effectively prune exploration, the trajectory-so-far is periodically evaluated by a relabeling module and further exploration continues only if it can be assigned a meaningful language instruction. All components in \ours{} are implemented with the same zero-shot base LLM.}
    \label{fig:enter-label}
\end{figure}

Unfortunately, language models struggle with such grounded instruction following out-of-the-box because LMs do not know about the myriad and ever changing interaction possibilities of different websites. For instance, on a new e-commerce website, a zero-shot LM browser agent may struggle to make a return or change order details, without expensive test-time exploration. Even simple tasks like choosing a flight can involve different UI element such as directly entering airport codes or interacting with drop-down menus, and a zero-shot agent cannot know a priori the correct thing to do. 
The most common solution is to provide LM browser agents with knowledge about new web interfaces via expert demonstrations, that can either be used for in-context learning \citep{yao2022react} or supervised fine-tuning \citep{lai2024autowebglm, shen2024scribeagent}. These demonstrations are either fully provided by human experts \citep{sodhi2023heap, yao2022react} or consist of human-generated trajectories paired with model-generated instructions \citep{lai2024autowebglm}. However, collecting human demonstrations that cover each possible use case for every website is an unattractively large, never-ending task.  Thus, in this work, we propose a method for training LM browser agents in a \emph{completely unsupervised way}, via synthetic demonstrations derived from interaction. %

\begin{figure*}[t]
    \centering
    \includegraphics[width=\linewidth]{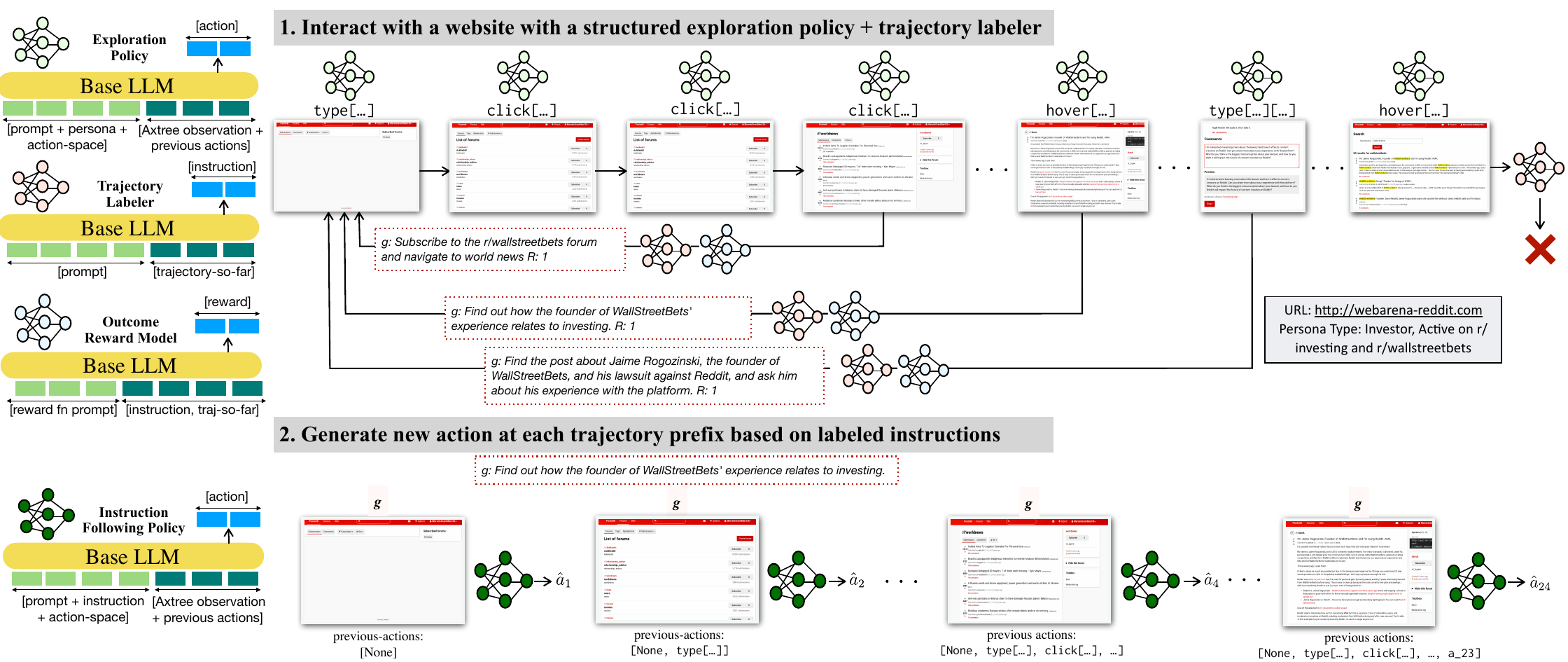}
    \caption{\textbf{Left:} \ours uses four components to interact with websites to create training examples, built out of zero-shot language models. \textbf{Right (Top):} An exploration episode on a website begins with sampling a persona, followed by generating persona-conditioned action sequences from the exploration policy. At fixed intervals, the trajectory labeler infers an instruction to describe the trajectory so far. If the resulting (instruction, trajectory) pair receives a low score from the ORM, the episode is pruned (indicated by a red cross). \textbf{Right (Bottom):} For each instruction, we retroactively generate a new action, given the (instruction, observation, previous actions) tuple to ensure that actions at each time-step correspond directly to the inferred instruction.}
    \label{fig:overview}
\end{figure*}

At a high level, our approach, \ours (Fig~\ref{fig:overview}), uses a language model exploration policy to perform extended interactions with a website, and another language model trajectory labeler to annotate trajectories with instructions.\footnote{Our code, data and trained models are available \href{https://nnetnav.dev/}{here}. } To effectively control the exponential space of meaningful interactions, \ours uses the hierarchical structure of language instructions as a pruning heuristic: for exploration to discover a meaningfully complex task, trajectory prefixes must correspond to meaningful sub-tasks. Thus, during an exploration episode, if a language model cannot label trajectory prefixes (at set time-steps) with a sub-task, further exploration is automatically pruned. Imposing such a structure over search not only enhances efficiency, but also results in complex and hierarchical instructions (See Table~\ref{tab:qual_examples} for examples). \ours prompts the same base language model for exploration, relabeling and inferring sub-tasks. 

We use \texttt{Llama-3.1-70B} \citep{dubey2024llama} to collect a large scale dataset of over 10k demonstrations (around 100k state, action transitions) from 20 websites, including 15 live, in-the-wild websites, and 5 self-hosted websites from WebArena \citep{zhou2023webarena}. We classify these instructions into various intents and find a highly diverse range of internet use cases, including \textit{flight booking}, \textit{finding recipes}, \textit{buying iPhones}, \textit{searching for trails}, \textit{commenting on github issues}, and \textit{posting on Reddit} (see Fig~\ref{fig:intent_burst_plot} for more examples). We use these demonstrations for supervised fine-tuning of \texttt{Llama-3.1-8B}. On WebArena, our model achieves a success rate of 16.3\%, outperforming zero-shot \texttt{GPT-4} by 2 points and reaching state-of-the-art performance among unsupervised methods. On WebVoyager \citep{he2024webvoyager}, our best model reaches a success rate of 35.2\%, outperforming zero-shot \texttt{GPT-4} by 1.7 points and all known open  methods on this task to the best of our knowledge. Interestingly, we find that \ours{} enables effective self-training---fine-tuning a smaller LM using \ours{} demonstrations generated by the same model yields a 4 point absolute improvement (from 1\% to 5\%) on WebArena. \ours{} opens up interesting avenues for open-ended discovery of workflows on unknown web-interfaces, without human supervision.

\section{Background}
\label{sec:background}
Following instructions on a web-browser is a multi-turn sequential decision making problem. Given an instruction $g$, a browser agent interacts with the browser by issuing a sequence of \emph{computer control} actions $\langle a_1, a_2, \ldots, a_T \rangle$ where each $a_i \in \mathcal{A}$ is drawn in response to an observation $o_i$. Executing an action causes a state transition based on some unknown environment dynamics, leading to a new observation $o_{i+1}$. The entire episode can be summarized as a \emph{trajectory} $\tau \coloneq \langle o_1, a_1, o_2, a_2, \ldots o_{T-1}, a_T, o_{T} \rangle$.  We formalize the instruction following agent as a mapping $\pi(a_t \mid o_t, \tau_{<t}; g)$ where  $\tau_{<t} \coloneq \langle o_1, a_1, \ldots a_{t-1} \rangle$ is the trajectory so far. In our case, observations are represented as either flattened DOM trees or website accessibility trees, and $\mathcal{A}$ consists of keyboard / mouse commands that operate on elements of these trees (see Appendix~\ref{sec:appendix_prompts} for the full action space).

\paragraph{LLMs for Browser Control.} Recent work explores using instruction-tuned large language models (LLMs) to directly parameterize the agent. These methods typically work in settings with textual observations and action spaces. At time-step $t$, the agent $\pi_\text{LM}$ is provided with the following context: the instruction $g$, the full action space described as a string, the current observation $o_t$,  and some representation of the trajectory-so-far $\tau_{<t}$, typically the action history. Given this information, the LLM generates an output that is parsed into an action. Typically, the LLM output contains both a reasoning step $r_t$ (e.g. \textit{Since my task is to buy a mug, given the current state, I should click on the buy now button}), and the chosen action command $a_t$ (e.g. \textit{click [1234]}).

Given expert demonstrations $\{g^i, \tau^i\}$ where $\tau^i \coloneq \langle o^i_1, r^i_1, a^i_1, o^i_2, r^i_2, a^i_2  \ldots o^i_T \rangle$, previous work adapts LM agents using demonstrations as in-context examples \citep[among others]{yao2022react, shinn2023reflexion, sun2023adaplanner, kim2023language} or as training data for supervised fine-tuning \citep{furuta2023multimodal, lai2024autowebglm, lu2024weblinx, patel2024large}. For supervised fine-tuning of $\pi_{\text{LM}}$ on a dataset of demonstrations, we construct training instances $\{(g^i, \tau^i_{<t}, o^i_t), (r^i_t, a^i_t)$\} where $r^i_t, a^i_t$ serves as the target reasoning step and action for an intermediate context $(g^i, \tau^i_{<t})$. 

\paragraph{Prior Methods for Synthetic Demonstrations.} Since collecting human demonstrations for browser agents is time consuming and costly, recent work uses synthetic demonstrations as training data  \citep{lai2024autowebglm, furuta2023multimodal, murty24bagel}. These methods start by sampling synthetic instructions from an instruction generator (a prompted LM that takes the website landing page and an optional user persona), and then use a zero-shot browser agent to convert these instructions into trajectories. Resulting demonstrations are filtered using either the ground truth reward function \citep{furuta2023multimodal}, or using another LM outcome reward function \citep{lai2024autowebglm, murty24bagel}. Most of these methods fine-tune smaller LMs using synthetic demonstrations from larger LMs.

Such \emph{instruction-first} methods for data collection face several challenges. First, synthetic instructions in these demonstrations are sampled from an ungrounded LM prior that generates only plausible\footnote{
We use the term \emph{plausible} for instructions that match a website's genre or intended use. For example, searching for clothes on a retail site or checking notifications on a social media platform. Not all plausible instructions are feasible.} instructions without ensuring feasibility; \eg, an instruction such as \textit{Delete the first post on r/callofdutyfans} for reddit is plausible, but not always feasible. Second, generated instructions are limited to those that reference visible features of the website; \eg, given the landing page of a github-like platform, no LM prior can generate instructions like \textit{Find information about Eric Bailey's contributions to the byteblaze project}, which require knowing about deeply embedded website-specific entities like \textit{Eric Bailey}. Finally, these methods provide no control over the complexity of instructions, and rely entirely on the LM or bespoke prompts to generate complex instructions.

\section{Our Approach}

Instead of starting with a sampled instruction, we start by sampling an \emph{interaction} first, and then retroactively labeling it into an instruction that is feasible by definition. \ours (Fig~\ref{fig:overview}) is an \emph{interaction-first} method for constructing demonstrations: An exploration policy interacts with a browser in a structured manner to sample long trajectories which are retroactively labeled into instructions (\S\ref{sec:exploration}). We then post-process each trajectory to add post-hoc actions corresponding to the generated instructions.

\subsection{LM Components}
\label{sec:components}
All components in \ours are implemented with a zero-shot instruction-tuned LLM, with different prompts (see Appendix~\ref{sec:appendix_prompts} for prompts).

\paragraph{Exploration Policy.} To interact with the environment, we use an exploration policy $\exploref{}$, implemented as -a prompted language model, similar to $\pi_\text{LM}$. Additionally, to simulate a diverse set of behaviors from users and improve the diversity of resulting trajectories, we seed each episode with a string description of a plausible user persona for the given website \citep{shanahan2023role, argyle2023out}. At each time-step, $\pi_\text{explore}$ is provided with the following context: a user persona, the list of available actions, the current observation $o_t$, and the action history. The output of $\pi_\text{explore}$ is then parsed into an action.

\paragraph{Summarizing Trajectory changes.} Actions issued by $\exploref{}$ result in a new observation in the environment. We summarize this change as a short string description via another module $\deltaf{}$, implemented using a language model. In particular, for any state $o_t$, action $a_t$ and the resulting next state $o_{t+1}$, $\delta_t = \deltaf{(o_t, a_t, o_{t+1})}$ produces a string-valued description of the changes in the observation as a result of the action. For a trajectory $\tau$, we denote the sequence of state changes as $\delta_\tau$.

\paragraph{Trajectory Labeler.} Given $\delta_\tau$, the trajectory labeler $\labelf{}$ produces a plausible instruction $\hat{g} = \labelf{(\delta_\tau)}$ that the agent could have followed to produce the given interaction. 

\paragraph{Outcome Reward Model.} Given $\hat{g}$ and $\delta_\tau$, the outcome reward model (ORM)  assigns a reward $\rewardf{(\hat{g}, \delta_\tau)} \in \{0, 1\}$, based on the degree to which state changes correspond to the given instruction $\hat{g}$.

\subsection{Sampling Demonstrations via Interactions}
\label{sec:exploration}

At specific time steps \( t \in \{t_1, t_2, \dots, t_\text{max} \} \), we apply a pruning heuristic to retroactively label the current trajectory. Given a partial trajectory \( \tau_{<t} \) after interacting with the environment for \( t \) steps, we compute a sub-task annotation \( \hat{g} = \labelf(\delta_{\tau_{<t}}) \). If this sub-task receives no reward, i.e., \( \rewardf(\hat{g}, \delta_{\tau_{<t}}) = 0 \), we prune the episode and sample a new rollout. Otherwise, we store \( (\hat{g}, \tau_{<t}) \) as a synthetic demonstration and continue exploration. Each episode typically generates multiple such demonstrations.

\paragraph{Post-processing with an Agent Policy.} Actions at each time-step in our our demonstration set are a result of un-directed exploration, and therefore might not be optimal for the retroactively generated instruction. Thus, we post-hoc annotate each state with a new action that directly corresponds to the generated instruction. Concretely, given every $(\hat{g}, o_i, \tau_{<t})$ tuple in our synthetic demonstration set, we use $\pi_\text{LM}$ to output a suitable action $\hat{a}_i$ given the instruction $\hat{g}$ and current observation $o_i$.

\paragraph{BofK sampling (Optional).} To further boost the quality of trajectories, we optionally use best-of-K (BofK) sampling. In particular, given \ours{} generated instructions, we sample $K-1$ additional trajectories, with $\pi_\text{LM}$ using the same base LLM. Then, for each instruction, we use our ORM to score each of the $K-1$ trajectories and the original trajectory, and pair the best trajectory with the given instruction, breaking ties arbitrarily.

\section{Main Experiments}
\subsection{Collecting Demonstrations in the Wild}
\label{sec:nnetnav_details}

We apply \ours{} on 20 websites to collect a dataset of over 10,000 demonstrations. We consider 15 live websites (same set as \citealt{he2024webvoyager}): Allrecipes, Amazon, Apple, ArXiv, BBC News, Booking, Cambridge Dictionary, Coursera, ESPN, GitHub, Google Flights, Google Map, Google Search, Huggingface, and Wolfram Alpha, and 5 self-hosted websites from WebArena (WA; \citealp{zhou2023webarena}).

We use instruct-tuned \texttt{Llama-3.1-70b} as the base LLM for all components in \ours, with $t_\text{max}$ set to 40, running \ours pruning every 4 time-steps at \{4, 8, 12, 16, ..., 40\}.  Additionally, we perform BofK sampling with $K=3$, using $\pi_\text{LM}$ (with the same \llama base model). While we only consider text based browser agents in this work, we release both accessibility tree strings as well as browser screenshots at each time step, to support future work on multi-modal browser agents. 

\begin{table}[!h]
\centering
\small
\renewcommand{\arraystretch}{1.2}
\begin{tabular}{lcc}
\toprule
\textbf{Difficulty} & \textbf{\ours (WA)} & \textbf{\ours (Live)} \\
\hline
Easy      & 498  & 1448 \\
Medium   & 2532 & 2369 \\
Hard        & 1164 & 1204 \\
Very-Hard   & 501  & 556  \\ \midrule
Total & 4695 & 5577 \\
\bottomrule
\end{tabular}
\caption{We report the breakdown of \ours demonstrations into categories defined based on the number of actions in the trajectory.}
\label{tab:complexity}
\end{table}

\paragraph{Diversity and Complexity.} To evaluate diversity in resulting instructions, we cluster them by intent for each website. We obtain these intents through a two-step procedure---we input instructions for each website into \gpttl{}, prompting it to identify common intents, and then classify each instruction into one of these intents in a second forward pass. On average, we identify 21 intents per website for self-hosted websites and 25 for live websites. Analyzing the distribution of these intents, we observe an average perplexity (PPL) of 13.5 for self-hosted sites and 16.2 for live websites. Higher perplexity suggests a more evenly distributed set of intents, indicating substantial diversity in the collected demonstrations. We provide a visual representation of this distribution as a sunburst plot in Appendix~\ref{sec:appendix_diversity}. 

To analyze the complexity of demonstrations, we categorize each demonstration into one of four levels based on the number of action sequences: \textit{easy} (fewer than 5 actions), \textit{medium} (5 to 10 actions), \textit{hard} (10 to 20 actions), and \textit{very hard} (over 20 actions). Table~\ref{tab:complexity} presents the distribution of demonstrations across these categories, 
showing a substantial number of complex demonstrations.

\begin{table*}[!t]
\centering
\small
\renewcommand{\arraystretch}{1.2}
\begin{tabular}{@{}lcccc@{}}
\toprule
\textbf{Agent} & \textbf{\#Params} & \textbf{WebArena SR} & \textbf{WebVoyager SR} & \textbf{Human Supervision  Used?}\\
\midrule
\hrow \multicolumn{5}{c}{\it Using Closed Models} \\

GPT-4 \citep{zhou2023webarena, he2024webvoyager} & Unknown  & 14.1 & 33.5 & \xmark\\
GPT-4-AWM \citep{wang2024agent} & Unknown & 35.5 & - & \xmark\\
GPT-4 + LLama-70b \citep{shen2024scribeagent} & Unknown & 50.0 & - & \cmark\\
\hrow \multicolumn{5}{c}{\it Using Open Models} \\
\llamas & 8B & 1.0 & 4.4 & \xmark \\
\citet{lai2024autowebglm} & 7B & 2.5 & - & \xmark \\
\citet{ou2024synatra} & 7B & 6.3 & - & \xmark \\ 
\citet{patel2024large} & 72B & 9.4 & - &  \xmark \\ 

LLaVa-7B PAE + Claude \citep{zhou2024proposer} & 7B & - & 22.3 & \xmark \\
LLaVa-34B PAE + Claude \citep{zhou2024proposer} & 34B & - & 33.0 & \xmark \\
Qwen2.5-7B-AgentTrek \citep{xu2024agenttrek} & 7B & 10.5 & - & \xmark \\
Qwen2.5-32B-AgentTrek \citep{xu2024agenttrek} & 32B & 16.3 & - & \xmark \\ 
\lourswa \textbf{(Ours)}  & 8B & \textbf{16.3} & 28.1 & \xmark \\
\loursow \textbf{(Ours)} & 8B  & 9.5 & \textbf{35.2} & \xmark \\
\loursb \textbf{(Ours)} & 8B & 14.9 & 34.1 & \xmark \\
\bottomrule
\end{tabular}
\caption{We present average success rate (SR) on browser tasks from WebArena and WebVoyager for various approaches, along with key details such as model size, the use of open LLMs and human supervision. For \citet{lai2024autowebglm}, we report results from the setting that does not use human supervision. Zero-shot \texttt{GPT-4} results are sourced from \citet{zhou2023webarena} and \citet{he2024webvoyager}. The last three rows report the performance of our fine-tuned \texttt{Llama-3.1-8b} agents, which achieve state-of-the-art results, outperforming zero-shot \texttt{GPT-4} and outperforming or matching prior open-model approaches with significantly fewer parameters, across both benchmarks.}
\label{tab:nnetnav6k_results}
\end{table*}

\subsection{Finetuning: Details and Results} 

We perform supervised fine-tuning of the smaller instruct-tuned \texttt{Llama-3.1-8B} with \ours{} demonstrations. To measure transfer between knowledge learned from live websites and self-hosted WebArena websites, we fine-tune on: only WebArena websites (\lourswa), only live websites (\loursow), and all websites together (\loursb).

As described in Section 2, each demonstration expands into multiple training instances, resulting in a total of ~100k training examples for the full dataset. We fine-tune for 2 epochs with a batch size of 128, truncating the max sequence length to 20000, with a learning rate of 2e-5, that is warmed with a linear scheduler over 500 gradient updates (more details can be found in Appendix~\ref{sec:training_details}). We use open-instruct \citep{wang2023far} for fine-tuning, and set up local inference servers using VLLM \citep{kwon2023efficient}. During inference, we sample with a temperature of 0.01 and perform nucleus sampling \citep{holtzman2019curious} with top-$p$ set to 0.9.

\paragraph{Benchmarks.} We evaluate models on 812 tasks from WebArena \citep{zhou2023webarena} and 557 tasks from WebVoyager \citep{he2024webvoyager}, omitting tasks in Google Flights and Booking, as they are no longer feasible (following \citealp{zhou2024proposer}). For WebArena, we report averaged success rate (SR) across all tasks based on the provided evaluator that measures functional correctness. For WebVoyager, we use the author-provided script that uses GPT-4V to judge success based on instructions and browser screenshots at each time step. We report the average across all websites.

\paragraph{Results.} We report our results in Table~\ref{tab:nnetnav6k_results}, where we present prior results from using closed models (typically \gpttl{}) as well as with open models. On WebArena, both \lourswa and \loursb outperform zero-shot \gpttl{}, with our best model achieving state-of-the-art performance among unsupervised methods.
On WebVoyager, \loursow and \loursb surpass zero-shot \gpttl{}, establishing a new state-of-the-art among open-source methods. Notably, they outperform the previous best OSS result from \citet{zhou2024proposer}, which relied on a significantly larger 34B-parameter vision-language model (VLM) and a closed-model verifier. Interestingly, we find that \lourswa, which is trained exclusively on WebArena websites, exhibits poor transfer to live websites. We analyze cross-website transfer next.

\subsection{Cross-Website Transfer}

We present per-website success rates of our fine-tuned models across all 18 websites in Table~\ref{tab:cross_website_transfer}.
For WebArena websites, by comparing columns 2 and 3, we find that 3 out of 5 websites benefit from incorporating in-domain data. By comparing columns 1 and 3, we observe an average performance drop of 1.8 points, with the most significant decrease on the \textit{Maps} domain. This decline is likely due to the semantic search capabilities in \textit{Google Maps}, which are absent in WebArena \textit{Maps}, necessitating more complex query formulation. For live websites, fine-tuning on in-domain live website data improves performance on 10 out of 13 domains, as indicated by comparing columns 1 and 3. The effect of incorporating out-of-domain WebArena data, however, is mixed. While it results in negative transfer for 7 websites and positive transfer for 6, the overall average performance decreases by 1.3 points. Notable gains are observed in \textit{ESPN}, \textit{Apple}, and \textit{GitHub}, suggesting potential synergies when fine-tuning on closely related domains.

Overall, fine-tuning with in-domain website data improves performance on 13 out of 18 websites. These findings underscore the importance of learning from unsupervised interaction on real websites, as relying solely on human-labeled trajectories from a limited set of simulated websites may be insufficient for developing generalist web agents.

\begin{table}[t]
\centering
\scriptsize 
\setlength{\tabcolsep}{3pt} 
\renewcommand{\arraystretch}{1.2} 
\resizebox{\columnwidth}{!}{ 
\begin{tabular}{@{}lccc@{}}
\toprule
\textbf{Website} & \textbf{\ours (WA)} & \textbf{\ours (Live)} & \textbf{\ours (Live+WA)}\\ 
\midrule
\hrow \multicolumn{4}{c}{\it Self-hosted Websites (WebArena)} \\
Reddit & \textbf{26.3} & 9.6 & 25.4 \\
Gitlab & \textbf{18.4} & 5.6 & 16.8 \\
Maps & \textbf{15.6} & 14.8 & 10.9 \\
CMS & \textbf{11.5} & 5.5 & 9.9 \\
Shopping & \textbf{13.0} & 9.9 & \textbf{13.0} \\
\hrow \multicolumn{4}{c}{\it Live Websites (WebVoyager)} \\
Allrecipes & 26.7 & \textbf{37.8} & 29.5 \\
Amazon & 24.4 & \textbf{43.9} & 34.1 \\
Apple & 32.6 & 27.9 & \textbf{34.9} \\
ArXiV & 27.9 & \textbf{46.5} & 44.2 \\
BBC News & 33.3 & \textbf{42.9} & 28.6 \\
Cambridge Dictionary & 46.5 & \textbf{58.1} & 48.8 \\
Coursera & \textbf{47.6} & 45.2 & 42.9 \\
ESPN & 20.5 & 22.7 & \textbf{27.3} \\
GitHub & 12.2 & 17.1 & \textbf{19.5} \\
Google Maps & 34.1 & \textbf{46.3} & 43.9 \\
Google Search & 0.0 & 2.7 & \textbf{6.2} \\
Huggingface & \textbf{30.2} & 18.6 & \textbf{30.2} \\
Wolfram Alpha & 26.1 & 43.5 & \textbf{45.7} \\
\bottomrule
\end{tabular}
}
\caption{Per-website success rates on all websites, using a \texttt{Llama-3.1-8b} agent fine-tuned on (1) the WebArena subset of \ours, (2) the live website subset of \ours, and (3) all demonstrations. On WebArena, incorporating in-domain data improves performance on 3 out of 5 websites (comparing columns 2 and 3). For live websites, incorporating in-domain data improves performance for 10 out of 13 websites (comparing columns 1 and 3). These results highlight the importance of scalable  methods to enable training on diverse websites.}
\label{tab:cross_website_transfer}
\vspace{-10pt}
\end{table}

\subsection{Error Analysis}
We analyze failure modes of our best models by manually annotating failed trajectories and categorizing them into fine-grained failure types. Specifically, we select 4 to 6 failure cases per website, prioritizing trajectories that exhibit distinct error patterns. Each failed trajectory is manually annotated with free-form natural language comments describing the nature of the failure \eg \textit{correctly navigates to cornell (maintainer of arxiv), but couldn't navigate to the section where there is info about undergraduate enrollment}. To categorize these failure cases, we use \texttt{GPT-4} to cluster similar annotations into distinct failure attributes. We then manually refine these attributes by merging redundant categories. Next, we prompt \texttt{GPT-4} to assign a score of \{-1, 0, 1\} to each trajectory based on these fine-grained attributes, where a score of 0 indicates \textit{not applicable}, 1 indicates \textit{positive reward} and -1 indicates \textit{negative reward}.

\begin{figure}
    \centering
    \includegraphics[width=0.8\linewidth]{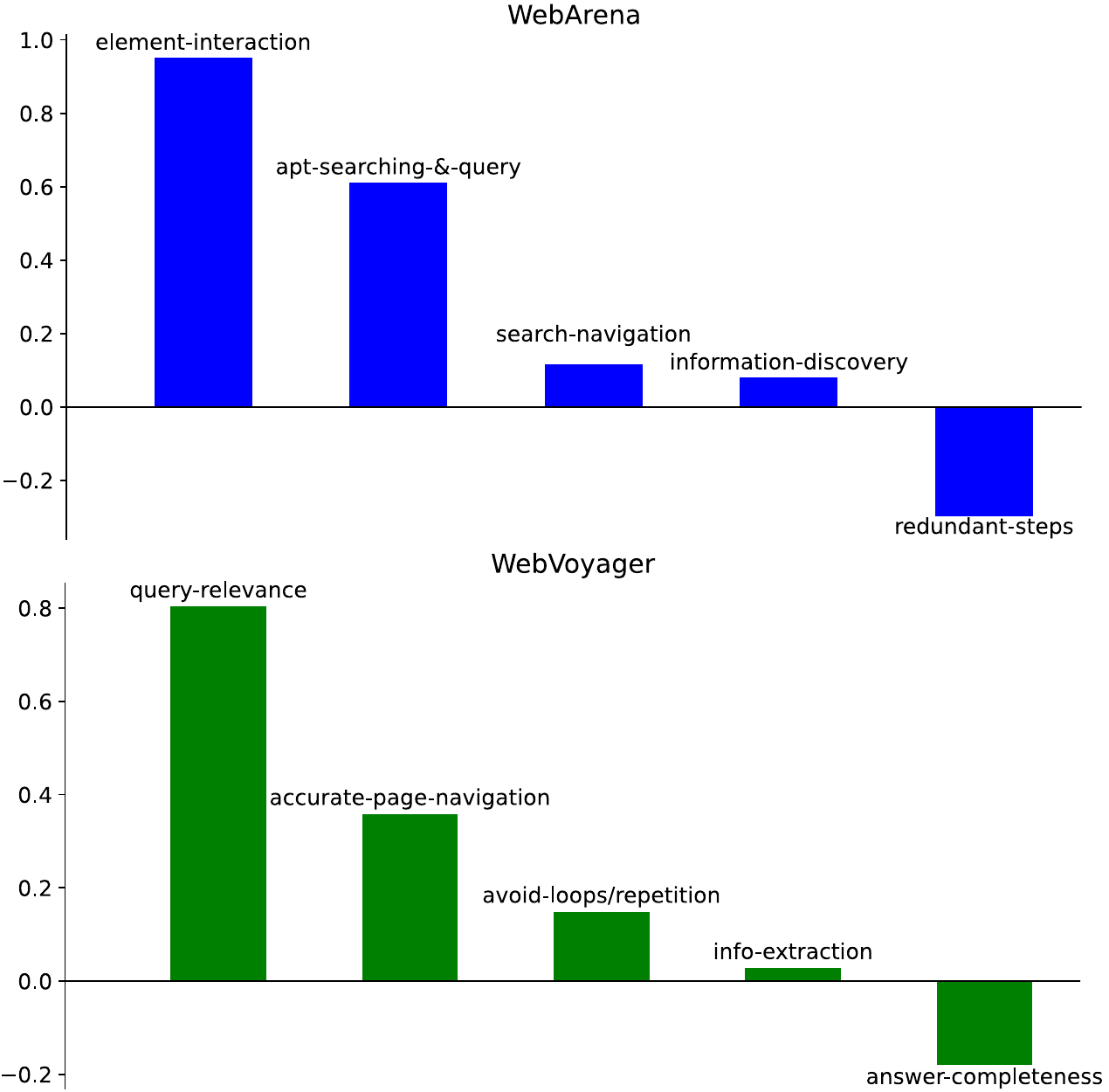}
    \caption{We score trajectories produced by our best models along 5 fine-grained attributes that were obtained by categorizing manually writing comments on model failures, for both WebArena and WebVoyager. We find that models can accurately interact with web-elements and conduct accurate searching and querying, however avoiding loops and redundant steps remains a challenge, as well as discovering information on web-sites by issuing multiple queries and piecing together information.}
    \label{fig:qual_analysis}
\end{figure}

Figure~\ref{fig:qual_analysis} presents the average reward per attribute for WebArena and WebVoyager. We observe that while our agents exhibit strong performance in element interaction and search functionality, they struggle with navigation efficiency, and sometimes execute redundant steps before reaching the target. Additionally, extracting information from structured data (\eg{}, tables on ESPN.com) remains a significant challenge. These findings suggest that future improvements should focus on heuristics to minimize unnecessary actions and enhancing the model’s ability to parse and retrieve structured web content.

\section{Controlled Experiments}
\label{sec:experiments}

We conduct controlled experiments on a smaller scale to compare \ours with baselines. In addition to evaluating on WebArena, we also consider MiniWoB++ \citep{shi2017world, liu2018reinforcement}. MiniWoB++ is a dataset of synthetic web-interfaces with a shared action space. Tasks on MiniWoB++ range from clicking on buttons to complex tasks like making a booking on a website. We use a subset of 8 complex tasks from MiniWoB++ as a toy benchmark to evaluate our method. We use the \texttt{bid}-based action space from BrowserGym \citep{drouin2024workarena}, consisting of 12 actions, and a DOM based observation space. Due to its synthetic nature, MiniWoB++ comes with an automatic reward function. We report the mean reward over 20 random seeds for each task, similar to \citet{drouin2024workarena}.

\subsection{Experimental Settings}

As before, we evaluate a \llamas based browser agent under the following settings:

\begin{enumerate}
    \item \textbf{Zero-Shot:} A baseline zero-shot agent, prompted using chain-of-thought prompting \citep{wei2022cot}. Next, we consider various fine-tuned models.
    \item \textbf{SFT (Instruction-First):} Supervised fine-tuning  of the \llamas agent using data collected via instruction-first sampling. Here, we use the same reward model for filtering demonstrations as \ours, and also sample the same number of demonstrations for fair comparison.
    \item \textbf{SFT (\ours):} Supervised fine-tuning of the \llamas agent with demonstrations collected via \ours.
    \item  \textbf{SFT (\ours + Distil.):} Ablation, where we only retain instructions found via \ours and re-generate trajectories by prompting the same large LM as an agent. We use this setting to isolate performance improvements attributable to \ours trajectories.
\end{enumerate}

For these small scale experiments, we use \texttt{gpt-4o-mini-2024-07-18} as the base LLM for both \ours and instruction-first methods.  For Instruction-first data collection, we sample 50 instructions per website for WebArena, and 80 instructions per interface in MiniWoB++, and prompt the instruction generator with the landing page as well as a persona (to improve diversity). For \ours, we use our exploration policy to generate 50 episodes per website for WebArena, and 80 episodes per interface for MiniWoB++. We set $T_\text{max}$ to 40 for WebArena, and 20 for MiniWoB++. For both MiniWoB++ and WebArena, we apply the pruning function every 4 time-steps. We use 16 persona types per website for WebArena, and 10 persona types per web-interface for MiniWoB++.

\begin{table}[!ht]
\centering
\small
\renewcommand{\arraystretch}{1.2}
\begin{tabular}{@{}lcc@{}}
\toprule
\textbf{Model Setting} & \textbf{MiniWoB++} & \textbf{WebArena} \\
\midrule
Zero-Shot                & 0.28             & 1.0              \\
SFT (Instruction-First)  & 0.28             & 4.2              \\
SFT (\ours)              & \textbf{0.48}    & \textbf{7.2}     \\
SFT (\ours + Distil.)    & 0.36             & 6.0              \\
\bottomrule
\end{tabular}
\caption{Controlled evaluation of \ours{} with instruction-first methods. We present results for MiniWoB++ and WebArena, averaged across domain, reporting mean reward for MiniWoB++ and task success rate (SR) for WebArena. Fine-tuning with \ours leads to the largest improvements: from 28\% to 48\% on MiniWoB++; from 1\% to 7.2\% on WebArena.}
\label{tab:main_results}
\end{table}

\subsection{Results}

\paragraph{\ours{} outperforms instruction-first methods.} We report results from all settings in Table~\ref{tab:main_results}. Fine-tuning \llamas using synthetic demonstrations generated by \ours yields significant improvements: an increase of 20 points on MiniWoB++ and over 6 points on WebArena. Notably, \ours outperforms instruction-first methods by a substantial margin, with gains of 12 points on MiniWoB++ and 1.2 points on WebArena. Interestingly, SFT (\ours) outperforms SFT (\ours + Distil.) on both MiniWoB++ and WebArena. This difference likely stems from the distinct procedures used to generate trajectories. In \ours, the model first acts, and the corresponding instruction is inferred afterward through a hindsight procedure. In contrast, \ours + Distil. provides the instruction upfront, sampling the trajectory later.

\begin{table} 
\captionsetup{type=table}
\centering
\renewcommand{\arraystretch}{1.1}
\small
\begin{tabular}{@{}lcc@{}}
\toprule
\textbf{Domain} & \textbf{Zero-Shot} & \textbf{Self-Train (\ours)} \\ \midrule
Shopping & 3.8 & \textbf{15.4} \\
CMS & 0.0 & 0.0 \\
Reddit & 0.0 & 0.0 \\
Gitlab & 0.0 & 0.0 \\
Maps & 0.0 & \textbf{7.1} \\ \midrule
Avg. & 1.0 & \textbf{5.3} \\  \bottomrule
\end{tabular}
\captionof{table}{We generate \ours demonstrations using \llamas, which we use for supervised fine-tuning of an agent based on the same LM, and find significant improvements on WebArena from 1\% to 5.3\%.}
\label{tab:selftrain}
\end{table}

\begin{figure}[!ht]
    \centering
\includegraphics[width=0.7\linewidth]{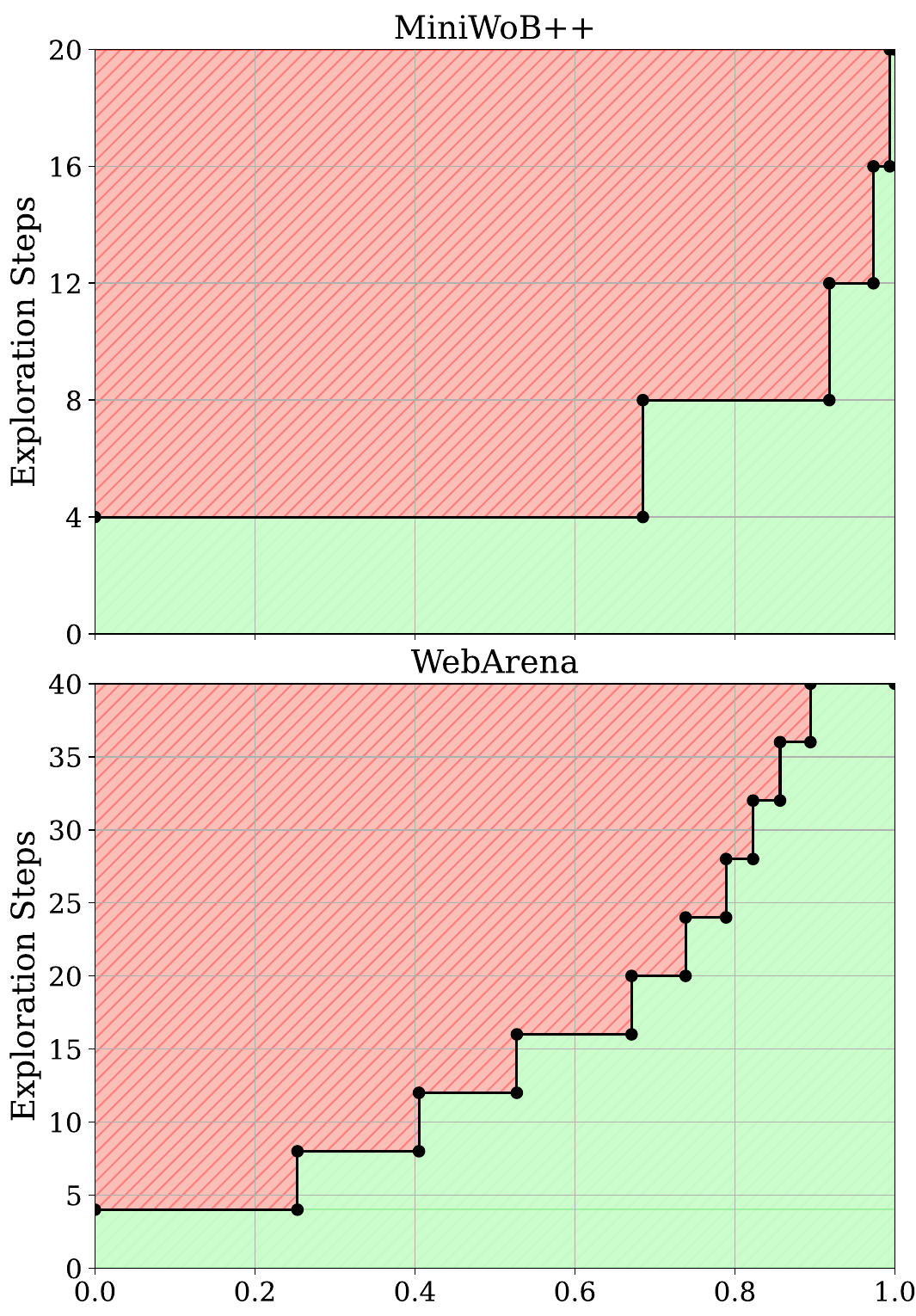}
  \caption{Horizontal lines indicate fraction of episodes terminating at corresponding y-axis exploration step. The red shaded area represents prevented actions, showing significant savings on both datasets.} 
  \label{fig:exploration_savings}
\end{figure}

\paragraph{Computational savings from \ours pruning.} We visualize overall improvements in exploration efficiency in Fig~\ref{fig:exploration_savings}. Each horizontal line depicts the fraction of interaction episodes that terminate at a specific time-step (indicated by the y-axis), with the red shaded area depicting additional actions that were prevented from early pruning. We find clear evidence of computational savings. In particular, over 60\% of all exploration episodes were pruned after 16 actions for WebArena. For MiniWoB++, 65\% of episodes were pruned after just 4 actions in MiniWoB++, which we identify as interactions where these first actions resulted in execution failures that our pruning heuristic successfully identified.

\paragraph{Self-training with \ours.} Can \ours demonstrations from an LM be used for improving the \emph{same} LM agent? To answer this, we collect another set of \ours demonstrations on WebArena, using \llamas as the base LM for data collection. Given the limitations of this smaller model, we anticipate fewer meaningful interactions. To compensate, we increase the number of episodes to 200 episodes per website, resulting in 302 demonstrations which we use for fine-tuning the same \llamas agent. From results in Table~\ref{tab:selftrain}, we find improvements of 4.3 points on WebArena.

\section{Related Work}
\paragraph{Language Conditioned Digital Assistants.}
Mapping instructions to actions in digital environments has been a long-standing goal in natural language understanding \citep{allen2007agents, branavan2009reinforcement}. Most pre-LLM approaches for this rely on 
expert demonstrations for behavioral cloning \citep{chen2011grounding, humphreys2022data}, along with appropriately shaped reward functions \citep[among others]{branavan2009reinforcement, liu2018reinforcement, misra2017mapping}. Here, learning is driven purely by synthetic demonstrations derived via (language model) exploration of websites.
\paragraph{Linguistic Priors for Exploration.}
Several prior works have used natural language priors to inform exploration for sequential decision making. 
\citet{harrison2017guiding} use a \emph{trained model} of associations between language and state/action pairs to guide exploration during policy learning. \citet{mu2022improving} use language annotations of states to train a goal generator module that provides intrinsic rewards for achieving generated goals. Similarly, \citet{du2023guiding} constrain exploration towards goals generated by a pre-trained LLM at each intermediate state of an agent. In constrast, \ours biases exploration through two news ways of using language priors. First, we use natural language as a way to filter meaningful interactions. Second, we use it as a pruning heuristic to navigate the potentially exponential search space of these interactions.
\paragraph{Training Data for LLM browser agents.} LLMs have shown strong performance over a wide range of language understanding tasks, and are increasingly being used to interpret language in grounded contexts such as browsers \citep[among others]{yao2022react, lai2024autowebglm, wang2024agent, patel2024large, lu2024weblinx}. Many of these approaches rely on human demonstrations, either for in-context learning \citep{yao2022react, sodhi2023heap, kim2023language} or for finetuning \citep{lu2024weblinx, shen2024scribeagent}. Since human demonstrations are costly, 
recent work trains LLM agents through synthetic demonstrations generated using instruction-first methods \citep{lai2024autowebglm, patel2024large}. One exception is \citet{murty24bagel}, which introduces an interaction-first method for generating synthetic demonstrations for in-context learning. Despite its novelty, their approach does not scale well to real websites due to the lack of a mechanism for effective exploration in environments with many possible interactions. In contrast, \ours also follows an interaction-first approach but improves efficiency by leveraging linguistically motivated pruning to navigate the space of meaningful interactions.

\section{Conclusion}
We propose \ours, a method for unsupervised interaction with websites ``in-the-wild'' that enables training browser agents with synthetic demonstrations. \ours flips the standard paradigm of synthetic data generation by first interacting with a website to generate trajectories and then hindsight relabeling trajectories into instructions. Real websites have a prohibitively large set of possible interactions; \ours searches over this space efficiently using a pruning function inspired by the hierarchical structure of language instructions: any complex instruction consists of language describable sub-tasks and so, if during an interaction a relabeling module cannot infer a meaningful sub-task for the trajectory-so-far, further exploration is pruned. We apply \ours to collect a diverse and complex set of 10k demonstrations from 15 live-websites and 5 self-hosted websites. We use these demonstrations for supervised finetuning of a small, \texttt{Llama-3.1-8b} model, achieving state-of-the-art results for unsupervised methods on both the WebArena and WebVoyager, surpassing zero-shot \texttt{GPT-4} by 1.7 to 2.2 points. \ours{} opens up the possibility of scaling up training data for generalist web agents across a broad range of web interfaces without any human intervention.

\section*{Acknowledgements}
CM is a fellow in the CIFAR Learning in Machines and Brains program. We thank Pratyusha Sharma, Allen Nie,  Moussa Doumbouya, Karel D'Oosterlinck, Anna Goldie and members of the Stanford NLP group for feedback on early drafts on the paper.

\section*{Impact Statement}

The deployment of unsupervised exploration with LLM agents on live websites has real-world implications, including website overload, unintended interactions, and the propagation of biases. To mitigate potential disruptions to websites, we constrain our agents to a maximum of 10 parallel instances, enforce a 0.5-second delay between actions, and prohibit login or content submission on live websites. We suggest that anyone using our work closely monitor these agents and establish robust monitoring frameworks to detect unintended behaviors and ensure compliance with ethical guidelines.  Additionally, training agents on \ours data from live websites can reinforce biases present in web content. We urge practitioners to conduct thorough bias audits before deployment.


\bibliography{icml2025}

\begin{thebibliography}{38}
\providecommand{\natexlab}[1]{#1}
\providecommand{\url}[1]{\texttt{#1}}
\expandafter\ifx\csname urlstyle\endcsname\relax
  \providecommand{\doi}[1]{doi: #1}\else
  \providecommand{\doi}{doi: \begingroup \urlstyle{rm}\Url}\fi

\bibitem[Allen et~al.(2007)Allen, Chambers, Ferguson, Galescu, Jung, Swift, and Taysom]{allen2007agents}
Allen, J., Chambers, N., Ferguson, G., Galescu, L., Jung, H., Swift, M., and Taysom, W.
\newblock Plow: a collaborative task learning agent.
\newblock In \emph{Proceedings of the 22nd National Conference on Artificial Intelligence - Volume 2}, AAAI'07, pp.\  1514–1519. AAAI Press, 2007.
\newblock ISBN 9781577353232.

\bibitem[Argyle et~al.(2023)Argyle, Busby, Fulda, Gubler, Rytting, and Wingate]{argyle2023out}
Argyle, L.~P., Busby, E.~C., Fulda, N., Gubler, J.~R., Rytting, C., and Wingate, D.
\newblock Out of one, many: Using language models to simulate human samples.
\newblock \emph{Political Analysis}, 31\penalty0 (3):\penalty0 337--351, 2023.

\bibitem[Branavan et~al.(2009)Branavan, Chen, Zettlemoyer, and Barzilay]{branavan2009reinforcement}
Branavan, S., Chen, H., Zettlemoyer, L., and Barzilay, R.
\newblock Reinforcement learning for mapping instructions to actions.
\newblock In Su, K.-Y., Su, J., Wiebe, J., and Li, H. (eds.), \emph{Proceedings of the Joint Conference of the 47th Annual Meeting of the {ACL} and the 4th International Joint Conference on Natural Language Processing of the {AFNLP}}, pp.\  82--90, Suntec, Singapore, August 2009. Association for Computational Linguistics.
\newblock URL \url{https://aclanthology.org/P09-1010}.

\bibitem[Chen \& Mooney(2011)Chen and Mooney]{chen2011grounding}
Chen, D. and Mooney, R.
\newblock Learning to interpret natural language navigation instructions from observations.
\newblock \emph{Proceedings of the AAAI Conference on Artificial Intelligence}, 25\penalty0 (1):\penalty0 859--865, Aug. 2011.
\newblock \doi{10.1609/aaai.v25i1.7974}.
\newblock URL \url{https://ojs.aaai.org/index.php/AAAI/article/view/7974}.

\bibitem[Drouin et~al.(2024)Drouin, Gasse, Caccia, Laradji, Del~Verme, Marty, Vazquez, Chapados, and Lacoste]{drouin2024workarena}
Drouin, A., Gasse, M., Caccia, M., Laradji, I.~H., Del~Verme, M., Marty, T., Vazquez, D., Chapados, N., and Lacoste, A.
\newblock {W}ork{A}rena: How capable are web agents at solving common knowledge work tasks?
\newblock In Salakhutdinov, R., Kolter, Z., Heller, K., Weller, A., Oliver, N., Scarlett, J., and Berkenkamp, F. (eds.), \emph{Proceedings of the 41st International Conference on Machine Learning}, volume 235 of \emph{Proceedings of Machine Learning Research}, pp.\  11642--11662. PMLR, 21--27 Jul 2024.
\newblock URL \url{https://proceedings.mlr.press/v235/drouin24a.html}.

\bibitem[Du et~al.(2023)Du, Watkins, Wang, Colas, Darrell, Abbeel, Gupta, and Andreas]{du2023guiding}
Du, Y., Watkins, O., Wang, Z., Colas, C., Darrell, T., Abbeel, P., Gupta, A., and Andreas, J.
\newblock Guiding pretraining in reinforcement learning with large language models.
\newblock \emph{arXiv preprint arXiv:2302.06692}, 2023.

\bibitem[Dubey et~al.(2024)Dubey, Jauhri, Pandey, Kadian, Al-Dahle, Letman, Mathur, Schelten, Yang, Fan, et~al.]{dubey2024llama}
Dubey, A., Jauhri, A., Pandey, A., Kadian, A., Al-Dahle, A., Letman, A., Mathur, A., Schelten, A., Yang, A., Fan, A., et~al.
\newblock The {Llama} 3 herd of models.
\newblock \emph{arXiv preprint arXiv:2407.21783}, 2024.

\bibitem[Furuta et~al.(2023)Furuta, Nachum, Lee, Matsuo, Gu, and Gur]{furuta2023multimodal}
Furuta, H., Nachum, O., Lee, K.-H., Matsuo, Y., Gu, S.~S., and Gur, I.
\newblock Multimodal web navigation with instruction-finetuned foundation models.
\newblock \emph{arXiv preprint arXiv:2305.11854}, 2023.

\bibitem[Harrison et~al.(2017)Harrison, Ehsan, and Riedl]{harrison2017guiding}
Harrison, B., Ehsan, U., and Riedl, M.~O.
\newblock Guiding reinforcement learning exploration using natural language.
\newblock \emph{arXiv preprint arXiv:1707.08616}, 2017.

\bibitem[He et~al.(2024)He, Yao, Ma, Yu, Dai, Zhang, Lan, and Yu]{he2024webvoyager}
He, H., Yao, W., Ma, K., Yu, W., Dai, Y., Zhang, H., Lan, Z., and Yu, D.
\newblock Webvoyager: Building an end-to-end web agent with large multimodal models.
\newblock \emph{arXiv preprint arXiv:2401.13919}, 2024.

\bibitem[Holtzman et~al.(2019)Holtzman, Buys, Du, Forbes, and Choi]{holtzman2019curious}
Holtzman, A., Buys, J., Du, L., Forbes, M., and Choi, Y.
\newblock The curious case of neural text degeneration.
\newblock \emph{arXiv preprint arXiv:1904.09751}, 2019.

\bibitem[Humphreys et~al.(2022)Humphreys, Raposo, Pohlen, Thornton, Chhaparia, Muldal, Abramson, Georgiev, Santoro, and Lillicrap]{humphreys2022data}
Humphreys, P.~C., Raposo, D., Pohlen, T., Thornton, G., Chhaparia, R., Muldal, A., Abramson, J., Georgiev, P., Santoro, A., and Lillicrap, T.
\newblock A data-driven approach for learning to control computers.
\newblock In \emph{International Conference on Machine Learning}, pp.\  9466--9482. PMLR, 2022.

\bibitem[Kim et~al.(2023)Kim, Baldi, and McAleer]{kim2023language}
Kim, G., Baldi, P., and McAleer, S.
\newblock Language models can solve computer tasks.
\newblock \emph{arXiv preprint arXiv:2303.17491}, 2023.

\bibitem[Kwon et~al.(2023)Kwon, Li, Zhuang, Sheng, Zheng, Yu, Gonzalez, Zhang, and Stoica]{kwon2023efficient}
Kwon, W., Li, Z., Zhuang, S., Sheng, Y., Zheng, L., Yu, C.~H., Gonzalez, J.~E., Zhang, H., and Stoica, I.
\newblock Efficient memory management for large language model serving with pagedattention.
\newblock In \emph{Proceedings of the ACM SIGOPS 29th Symposium on Operating Systems Principles}, 2023.

\bibitem[Lai et~al.(2024)Lai, Liu, Iong, Yao, Chen, Shen, Yu, Zhang, Zhang, Dong, and Tang]{lai2024autowebglm}
Lai, H., Liu, X., Iong, I.~L., Yao, S., Chen, Y., Shen, P., Yu, H., Zhang, H., Zhang, X., Dong, Y., and Tang, J.
\newblock Autowebglm: Bootstrap and reinforce a large language model-based web navigating agent, 2024.

\bibitem[Liu et~al.(2018)Liu, Guu, Pasupat, Shi, and Liang]{liu2018reinforcement}
Liu, E.~Z., Guu, K., Pasupat, P., Shi, T., and Liang, P.
\newblock Reinforcement learning on web interfaces using workflow-guided exploration.
\newblock In \emph{International Conference on Learning Representations}, 2018.

\bibitem[L{\`u} et~al.(2024)L{\`u}, Kasner, and Reddy]{lu2024weblinx}
L{\`u}, X.~H., Kasner, Z., and Reddy, S.
\newblock Weblinx: Real-world website navigation with multi-turn dialogue.
\newblock \emph{arXiv preprint arXiv:2402.05930}, 2024.

\bibitem[Misra et~al.(2017)Misra, Langford, and Artzi]{misra2017mapping}
Misra, D., Langford, J., and Artzi, Y.
\newblock Mapping instructions and visual observations to actions with reinforcement learning.
\newblock \emph{arXiv preprint arXiv:1704.08795}, 2017.

\bibitem[Mu et~al.(2022)Mu, Zhong, Raileanu, Jiang, Goodman, Rockt{\"a}schel, and Grefenstette]{mu2022improving}
Mu, J., Zhong, V., Raileanu, R., Jiang, M., Goodman, N., Rockt{\"a}schel, T., and Grefenstette, E.
\newblock Improving intrinsic exploration with language abstractions.
\newblock \emph{Advances in Neural Information Processing Systems}, 35:\penalty0 33947--33960, 2022.

\bibitem[Murty et~al.(2024)Murty, Manning, Shaw, Joshi, and Lee]{murty24bagel}
Murty, S., Manning, C.~D., Shaw, P., Joshi, M., and Lee, K.
\newblock {BAGEL}: Bootstrapping agents by guiding exploration with language.
\newblock In Salakhutdinov, R., Kolter, Z., Heller, K., Weller, A., Oliver, N., Scarlett, J., and Berkenkamp, F. (eds.), \emph{Proceedings of the 41st International Conference on Machine Learning}, volume 235 of \emph{Proceedings of Machine Learning Research}, pp.\  36894--36910. PMLR, 21--27 Jul 2024.
\newblock URL \url{https://proceedings.mlr.press/v235/murty24a.html}.

\bibitem[Ou et~al.(2024)Ou, Xu, Madaan, Liu, Lo, Sridhar, Sengupta, Roth, Neubig, and Zhou]{ou2024synatra}
Ou, T., Xu, F.~F., Madaan, A., Liu, J., Lo, R., Sridhar, A., Sengupta, S., Roth, D., Neubig, G., and Zhou, S.
\newblock Synatra: Turning indirect knowledge into direct demonstrations for digital agents at scale.
\newblock \emph{arXiv preprint arXiv:2409.15637}, 2024.

\bibitem[Patel et~al.(2024)Patel, Hofmarcher, Leoveanu-Condrei, Dinu, Callison-Burch, and Hochreiter]{patel2024large}
Patel, A., Hofmarcher, M., Leoveanu-Condrei, C., Dinu, M.-C., Callison-Burch, C., and Hochreiter, S.
\newblock Large language models can self-improve at web agent tasks.
\newblock \emph{arXiv preprint arXiv:2405.20309}, 2024.

\bibitem[Rajbhandari et~al.(2020)Rajbhandari, Rasley, Ruwase, and He]{rajbhandari2020zero}
Rajbhandari, S., Rasley, J., Ruwase, O., and He, Y.
\newblock Zero: Memory optimizations toward training trillion parameter models.
\newblock In \emph{SC20: International Conference for High Performance Computing, Networking, Storage and Analysis}, pp.\  1--16. IEEE, 2020.

\bibitem[Shanahan et~al.(2023)Shanahan, McDonell, and Reynolds]{shanahan2023role}
Shanahan, M., McDonell, K., and Reynolds, L.
\newblock Role play with large language models.
\newblock \emph{Nature}, 623\penalty0 (7987):\penalty0 493--498, 2023.

\bibitem[Shen et~al.(2024)Shen, Jain, Xiao, Amlekar, Hadji, Podolny, and Talwalkar]{shen2024scribeagent}
Shen, J., Jain, A., Xiao, Z., Amlekar, I., Hadji, M., Podolny, A., and Talwalkar, A.
\newblock Scribeagent: Towards specialized web agents using production-scale workflow data.
\newblock \emph{arXiv preprint arXiv:2411.15004}, 2024.

\bibitem[Shi et~al.(2017)Shi, Karpathy, Fan, Hernandez, and Liang]{shi2017world}
Shi, T., Karpathy, A., Fan, L., Hernandez, J., and Liang, P.
\newblock World of bits: An open-domain platform for web-based agents.
\newblock In \emph{International Conference on Machine Learning}, pp.\  3135--3144. PMLR, 2017.

\bibitem[Shinn et~al.(2023)Shinn, Labash, and Gopinath]{shinn2023reflexion}
Shinn, N., Labash, B., and Gopinath, A.
\newblock Reflexion: an autonomous agent with dynamic memory and self-reflection.
\newblock \emph{arXiv preprint arXiv:2303.11366}, 2023.

\bibitem[Sodhi et~al.(2023)Sodhi, Branavan, and McDonald]{sodhi2023heap}
Sodhi, P., Branavan, S., and McDonald, R.
\newblock Heap: Hierarchical policies for web actions using llms.
\newblock \emph{arXiv preprint arXiv:2310.03720}, 2023.

\bibitem[Sun et~al.(2023)Sun, Zhuang, Kong, Dai, and Zhang]{sun2023adaplanner}
Sun, H., Zhuang, Y., Kong, L., Dai, B., and Zhang, C.
\newblock Adaplanner: Adaptive planning from feedback with language models.
\newblock \emph{arXiv preprint arXiv:2305.16653}, 2023.

\bibitem[Wang et~al.(2023)Wang, Ivison, Dasigi, Hessel, Khot, Chandu, Wadden, MacMillan, Smith, Beltagy, and Hajishirzi]{wang2023far}
Wang, Y., Ivison, H., Dasigi, P., Hessel, J., Khot, T., Chandu, K.~R., Wadden, D., MacMillan, K., Smith, N.~A., Beltagy, I., and Hajishirzi, H.
\newblock How far can camels go? exploring the state of instruction tuning on open resources, 2023.

\bibitem[Wang et~al.(2024)Wang, Mao, Fried, and Neubig]{wang2024agent}
Wang, Z.~Z., Mao, J., Fried, D., and Neubig, G.
\newblock Agent workflow memory.
\newblock \emph{arXiv preprint arXiv:2409.07429}, 2024.

\bibitem[Wei et~al.(2022)Wei, Wang, Schuurmans, Bosma, ichter, Xia, Chi, Le, and Zhou]{wei2022cot}
Wei, J., Wang, X., Schuurmans, D., Bosma, M., ichter, b., Xia, F., Chi, E., Le, Q.~V., and Zhou, D.
\newblock Chain-of-thought prompting elicits reasoning in large language models.
\newblock In Koyejo, S., Mohamed, S., Agarwal, A., Belgrave, D., Cho, K., and Oh, A. (eds.), \emph{Advances in Neural Information Processing Systems}, volume~35, pp.\  24824--24837. Curran Associates, Inc., 2022.

\bibitem[Winograd(1972)]{winograd1972understand}
Winograd, T.
\newblock Understanding natural language.
\newblock \emph{Cognitive Psychology}, 3\penalty0 (1):\penalty0 1--191, 1972.
\newblock ISSN 0010-0285.
\newblock \doi{https://doi.org/10.1016/0010-0285(72)90002-3}.
\newblock URL \url{https://www.sciencedirect.com/science/article/pii/0010028572900023}.

\bibitem[Xie et~al.(2024)Xie, Zhang, Chen, Li, Zhao, Cao, Hua, Cheng, Shin, Lei, et~al.]{xie2024osworld}
Xie, T., Zhang, D., Chen, J., Li, X., Zhao, S., Cao, R., Hua, T.~J., Cheng, Z., Shin, D., Lei, F., et~al.
\newblock Osworld: Benchmarking multimodal agents for open-ended tasks in real computer environments.
\newblock \emph{arXiv preprint arXiv:2404.07972}, 2024.

\bibitem[Xu et~al.(2024)Xu, Lu, Shen, Wang, Wang, Mao, Xiong, and Yu]{xu2024agenttrek}
Xu, Y., Lu, D., Shen, Z., Wang, J., Wang, Z., Mao, Y., Xiong, C., and Yu, T.
\newblock Agenttrek: Agent trajectory synthesis via guiding replay with web tutorials.
\newblock \emph{arXiv preprint arXiv:2412.09605}, 2024.
\newblock URL \url{https://arxiv.org/abs/2412.09605}.

\bibitem[Yao et~al.(2022)Yao, Zhao, Yu, Du, Shafran, Narasimhan, and Cao]{yao2022react}
Yao, S., Zhao, J., Yu, D., Du, N., Shafran, I., Narasimhan, K.~R., and Cao, Y.
\newblock {ReAct}: Synergizing reasoning and acting in language models.
\newblock In \emph{The Eleventh International Conference on Learning Representations}, 2022.

\bibitem[Zhou et~al.(2023)Zhou, Xu, Zhu, Zhou, Lo, Sridhar, Cheng, Bisk, Fried, Alon, et~al.]{zhou2023webarena}
Zhou, S., Xu, F.~F., Zhu, H., Zhou, X., Lo, R., Sridhar, A., Cheng, X., Bisk, Y., Fried, D., Alon, U., et~al.
\newblock Webarena: A realistic web environment for building autonomous agents.
\newblock \emph{arXiv preprint arXiv:2307.13854}, 2023.

\bibitem[Zhou et~al.(2024)Zhou, Yang, Lin, Bai, Zhou, Wang, Levine, and Li]{zhou2024proposer}
Zhou, Y., Yang, Q., Lin, K., Bai, M., Zhou, X., Wang, Y.-X., Levine, S., and Li, E.
\newblock Proposer-agent-evaluator (pae): Autonomous skill discovery for foundation model internet agents.
\newblock \emph{arXiv preprint arXiv:2412.13194}, 2024.

\end{thebibliography}
\bibliographystyle{icml2025}

\newpage
\appendix
\onecolumn

\appendix
\section{Prompts for LM components}
\label{sec:appendix_prompts}

\subsection{MiniWoB++}
\label{sec:miniwob_prompts}
We start by presenting all prompts for MiniWoB++. The action space for MiniWob++ is:

\begin{lstlisting}[style=textFileStyle, caption=Action Space ]
noop(wait_ms: float = 1000)
    Examples:
        noop()
        noop(500)

scroll(delta_x: float, delta_y: float)
    Examples:
        scroll(0, 200)
        scroll(-50.2, -100.5)

fill(bid: str, value: str)
    Examples:
        fill('237', 'example value')
        fill('45', 'multi-line\nexample')
        fill('a12', 'example with "quotes"')

select_option(bid: str, options: str | list[str])
    Examples:
        select_option('a48', 'blue')
        select_option('c48', ['red', 'green', 'blue'])

click(bid: str, button: Literal['left', 'middle', 'right'] = 'left', modifiers: list[typing.Literal['Alt', 'Control', 'Meta', 'Shift']] = [])
    Examples:
        click('a51')
        click('b22', button='right')
        click('48', button='middle', modifiers=['Shift'])

dblclick(bid: str, button: Literal['left', 'middle', 'right'] = 'left', modifiers: list[typing.Literal['Alt', 'Control', 'Meta', 'Shift']] = [])
    Examples:
        dblclick('12')
        dblclick('ca42', button='right')
        dblclick('178', button='middle', modifiers=['Shift'])

hover(bid: str)
    Examples:
        hover('b8')

press(bid: str, key_comb: str)
    Examples:
        press('88', 'Backspace')
        press('a26', 'Control+a')
        press('a61', 'Meta+Shift+t')

focus(bid: str)
    Examples:
        focus('b455')

clear(bid: str)
    Examples:
        clear('996')

drag_and_drop(from_bid: str, to_bid: str)
    Examples:
        drag_and_drop('56', '498')

upload_file(bid: str, file: str | list[str])
    Examples:
        upload_file('572', 'my_receipt.pdf')
        upload_file('63', ['/home/bob/Documents/image.jpg', '/home/bob/Documents/file.zip'])

Only a single action can be provided at once. Example:
fill('a12', 'example with "quotes"')

If you are done exploring, you can issue the stop action: ```stop```

Here is an example with chain of thought of a valid action when clicking on a button: "In order to accomplish my goal I need to click on the button with bid 12. In summary, the next action I will perform is ```click("12")```
\end{lstlisting}
\vspace{2em}
This is then directly used for various prompts as \texttt{\{action\_str\}}.

\begin{lstlisting}[style=textFileStyle, caption=Prompt for the Exploration Policy $\pi_\text{explore}$]
You are an autonomous intelligent agent tasked with performing tasks on a web interface. Your objective is to simulate a task that a person might request, by interacting with the interface through the use of specific actions.

Here's the information you'll have:
DOM Representation: This is the current webpage's Document Object Model (DOM) representation as a string.
The previous action: This is the action you just performed. It may be helpful to track your progress.
Trajectory: This is a sequence of natural language descriptions of the agent's interaction with the web-browser.
Person Description: The description of a specific kind of person whose task you are supposed to simulate.

You can perform the following actions: {action_str}

To be successful, it is very important to follow the following rules:
1. You should only issue an action that is valid given the current observation.
2. You should only issue one action at a time.
3. You should reason step by step and then issue the next action.
4. Make sure to wrap your action in a code block using triple backticks.
5. The DOM / Accessibility Tree only shows the visible part of the webpage. If you need to interact with elements that are not visible, you can scroll to them using the scroll action. Often submit buttons are not visible and are at the bottom of the page. To scroll to the bottom of the page, use the scroll action with a large value for the y coordinate.
6. To generate an interesting task, make sure you issue atleast 4 actions before stopping. More interesting tasks typically involve more interactions with the browser.
7. You can issue atmost 20 actions before stopping, but feel free to output the stop action early if you want to stop exploring. Don't generate anything after stop.
\end{lstlisting}

\begin{lstlisting}[style=textFileStyle, caption=Prompt for $\deltaf{}$]
You are given the output of an action taken by an autonomous intelligent agent navigating a web-interface to fulfill a task given by a user.  Your objective is to produce a description of the changes made to the state of the browser.

Here's the information you'll have:
Initial state of the browser as a DOM representation: This is the webpage's Document Object Model (DOM) representation as a string.
Final state of the browser as a DOM representation: This is the DOM representation after the agent took the action.

The action taken by the agent: This is the action taken by the agent to change the state of the browser. 

The actions the agent can take come from the following categories: {action_str}

To be successful, it is very important to follow the following rules:
1. Explictly think about the various features on the website and how the interaction with the website changed these features
2. Provide the description of changes in one or two sentences.
3. Strictly follow the format "State change: <your-answer>" for your output
\end{lstlisting}

\begin{lstlisting}[style=textFileStyle, caption=Prompt for the Trajectory Labeling function $\labelf{}$]
Given a task from a user, an autonomous intelligent agent carries out a sequence of actions on a web-interface. 

The actions the agent can take fall under the following categories: {action_str}

Your objective is to guess the instruction the user gave, given the following information:
Trajectory: This is a sequence of natural language descriptions of the agent's interaction with the web-browser.

To be successful, it is very important to follow the following rules:
1. Explictly think about how the trajectory is a valid way to achieve the instruction, before outputing the instruction.
2. Start by thinking by outputing Thought: <your-reasoning>.
3. End your answer by strictly following the format "Instruction: <your-answer>" for your output.
\end{lstlisting}

\begin{lstlisting}[style=textFileStyle, caption=Prompt for the reward function $\rewardf{}$]
An autonomous intelligent agent navigating a web browser is given an instruction by a user. Your objective is to give a score to the agent based on how well it completed its task. Your score must be on the scale of 1 to 5. Give a score of 5 only when there are no errors. To do this task you are provided with the following information:

Instruction: This is the natural language instruction given to the agent.
Trajectory: This is a sequence of natural language descriptions of the agent's interaction with the web-browser.

To be successful, it is very important to follow the following rules:
1. Explictly think about what is needed to follow the instruction correctly on the website and if the trajectory reflects these steps.
2 Give a score of 4 if there are very minor errors, or if the task was more than 70% completed. Give a score of 3 (or below) if the model made very little progress towards the given instruction or if there are major errors.
3. Start by thinking by outputing Thought: <your-reasoning>.
4. End your answer by strictly following the format "Reward: <your-answer>" for your output
\end{lstlisting}

\begin{lstlisting}[style=textFileStyle, caption=Prompt for the base LLM agent $\policy$]
You are an autonomous intelligent agent tasked with performing tasks on a web interface. These tasks will be accomplished through the use of specific actions you can issue.

Here's the information you'll have:
DOM Representation: This is the current webpage's Document Object Model (DOM) representation as a string.
The user's objective: This is the task you're trying to complete.
The previous action: This is the action you just performed. It may be helpful to track your progress.

You can perform the following actions: {action_str}

To be successful, it is very important to follow the following rules:
1. You should only issue an action that is valid given the current observation
2. You should only issue one action at a time.
3. You should follow the examples to reason step by step and then issue the next action.
4. Make sure to wrap your action in a code block using triple backticks.
5. The DOM / Accessibility Tree only shows the visible part of the webpage. If you need to interact with elements that are not visible, you can scroll to them using the scroll action. Often submit buttons are not visible and are at the bottom of the page. To scroll to the bottom of the page, use the scroll action with a large value for the y coordinate.
6. Issue stop action when you think you have achieved the objective. Don't generate anything after stop.
\end{lstlisting}

\subsection{Prompts for WebArena and Live Websites}
\label{sec:webarena_prompts}

Next, we present all prompts for running policies on self-hosted WebArena websites and live websites. The action space is:

\begin{lstlisting}[style=textFileStyle, caption=Action Space ]
Page Operation Actions:
`click [id]`: This action clicks on an element with a specific id on the webpage.
`type [id] [content] [press_enter_after=0|1]`: Use this to type the content into the field with id. By default, the "Enter" key is pressed after typing unless press_enter_after is set to 0.
`hover [id]`: Hover over an element with id.
`press [key_comb]`:  Simulates the pressing of a key combination on the keyboard (e.g., Ctrl+v).
`scroll [direction=down|up]`: Scroll the page up or down.

Tab Management Actions:
`new_tab`: Open a new, empty browser tab.
`tab_focus [tab_index]`: Switch the browser's focus to a specific tab using its index.
`close_tab`: Close the currently active tab.

URL Navigation Actions:
`goto [url]`: Navigate to a specific URL.
`go_back`: Navigate to the previously viewed page.
`go_forward`: Navigate to the next page (if a previous 'go_back' action was performed).

Completion Action:
`stop ["done"]`: Issue this action when you are done.
\end{lstlisting}
\vspace{2em}
Additionally, for WebArena, models can visit the homepage at \texttt{http://homepage.com}, which lists all the websites on WebArena. This is then directly used for various prompts as \texttt{\{action\_str\}}.

\begin{lstlisting}[style=textFileStyle, caption=Prompt for the Exploration Policy $\pi_\text{explore}$ in WebArena]
You are an autonomous intelligent agent tasked with navigating a web browser.  Your objective is to simulate a task that a person might perform, by interacting with the browser through the use of specific actions.

Here's the information you'll have:

The current web page's accessibility tree: This is a simplified representation of the webpage, providing key information.
The current web page's URL: This is the page you're currently navigating.
The open tabs: These are the tabs you have open.
The previous action: This is the action you just performed. It may be helpful to track your progress.
Trajectory: This is a sequence of natural language descriptions of the agent's interaction with the web-browser.
Person Description: The description of a specific kind of person whose task you are supposed to simulate.

The actions you can perform fall into several categories: {action_str}

To be successful, it is very important to follow the following rules:
1. You should only issue an action that is valid given the current observation
2. You should only issue one action at a time.
3. You should follow the examples to reason step by step and then issue the next action.
4. Generate the action in the correct format. Start by reasoning out the current situation. End with "In summary, the next action I will perform is" phrase, followed by action inside ``````. For example, "Let's think step-by-step. Given the current state, I need to click on the like button which has id 1234. In summary, the next action I will perform is ```click [1234]```".
5. To generate an interesting task, make sure you issue atleast 4 actions before stopping. More interesting tasks typically involve more interactions with the browser.
6. You can issue atmost 40 actions before stopping, but feel free to output the stop action early if you want to stop exploring. Don't generate anything after stop.

Here are some example outputs for some random tasks:
1. Let's think step-by-step. This page list the information of HP Inkjet Fax Machine, which is the product identified in the objective. Its price is $279.49. I think I have achieved the objective. I will issue the stop action with the answer. In summary, the next action I will perform is ```stop [$279.49]```
2. Let's think step-by-step. This page has a search box whose ID is [164]. According to the nominatim rule of openstreetmap, I can search for the restaurants near a location by "restaurants near". I can submit my typing by pressing the Enter afterwards. In summary, the next action I will perform is ```type [164] [restaurants near CMU] [1]``
\end{lstlisting}

For Exploration on live websites, we add a few extra rules for our model to ensure safety and terminate exploration when CAPTCHAs or logins are triggered.

\begin{lstlisting}[style=textFileStyle, caption=Prompt for the Exploration Policy $\pi_\text{explore}$ in WebArena]
You are an autonomous intelligent agent tasked with navigating a web browser.  Your objective is to simulate a task that a person might perform, by interacting with the browser through the use of specific actions.

Here's the information you'll have:

The current web page's accessibility tree: This is a simplified representation of the webpage, providing key information.
The current web page's URL: This is the page you're currently navigating.
The open tabs: These are the tabs you have open.
The previous action: This is the action you just performed. It may be helpful to track your progress.
Trajectory: This is a sequence of natural language descriptions of the agent's interaction with the web-browser.
Person Description: The description of a specific kind of person whose task you are supposed to simulate.

The actions you can perform fall into several categories:

Page Operation Actions:
`click [id]`: This action clicks on an element with a specific id on the webpage.
`type [id] [content] [press_enter_after=0|1]`: Use this to type the content into the field with id. By default, the "Enter" key is pressed after typing unless press_enter_after is set to 0.
`hover [id]`: Hover over an element with id.
`press [key_comb]`:  Simulates the pressing of a key combination on the keyboard (e.g., Ctrl+v).
`scroll [direction=down|up]`: Scroll the page up or down.

Tab Management Actions:
`new_tab`: Open a new, empty browser tab.
`tab_focus [tab_index]`: Switch the browser's focus to a specific tab using its index.
`close_tab`: Close the currently active tab.

URL Navigation Actions:
`goto [url]`: Navigate to a specific URL.
`go_back`: Navigate to the previously viewed page.
`go_forward`: Navigate to the next page (if a previous 'go_back' action was performed).

Completion Action:
`stop ["done"]`: Issue this action when you are done. You can use the stop action to convey a message to the user, but know that your interaction will terminate after this.

Homepage:
If you want to visit other websites, check out the homepage at http://homepage.com. It has a list of websites you can visit.

To be successful, it is very important to follow the following rules:
1. You should only issue an action that is valid given the current observation
2. You should only issue one action at a time.
3. You should follow the examples to reason step by step and then issue the next action.
4. Generate the action in the correct format. Start with a "In summary, the next action I will perform is" phrase, followed by action inside ``````. For example, "In summary, the next action I will perform is ```click [1234]```".
5. To generate an interesting task, make sure you issue atleast 4 actions before stopping. More interesting tasks typically involve more interactions with the browser.
6. You can issue atmost 40 actions before stopping, but feel free to output the stop action early if you want to stop exploring. Don't generate anything after stop. 

Finally, here are some more rules that you should follow for specific websites:

1. On bookings and google flight, please use the date picker to choose start date (2025-01-01) and end date (2025-01-03). Make sure you click search after you input the dates.
2. Don't click disabled or invisible links on any website.
3. On google map, try to search for some locations around the world.
4. On all websites, don't click "Enroll", "Sign up", or other buttons indicating creating new accounts. Instead, just stop by issuing ```stop['exit']``` if you want to pass control to a user to sign-up.
5. On all websites, don't click "Sign in", "Log in through Google", or other buttons indicating logging into existing accounts. Instead, just stop if you want to pass control to a user to sign-in by issuing ```stop['exit']``` action.
6. On arxiv.org, please always check html version of the papers. Don't click view PDF.
7. When dealing pop ups, click "Maybe later" or other links that can turn off the pop up temporarily. 


Here are some example outputs for some random tasks:
1. Let's think step-by-step. This page list the information of HP Inkjet Fax Machine, which is the product identified in the objective. Its price is $279.49. I think I have achieved the objective. I will issue the stop action with the answer. In summary, the next action I will perform is ```stop [$279.49]```
2. Let's think step-by-step. This page has a search box whose ID is [164]. According to the nominatim rule of openstreetmap, I can search for the restaurants near a location by "restaurants near". I can submit my typing by pressing the Enter afterwards. In summary, the next action I will perform is ```type [164] [restaurants near CMU] [1]```
3. Let's think step-by-step. I want to see more of the page since the submit button is not visible. I will scroll down to see the submit button. In summary, the next action I will perform is ```scroll [down]```.
\end{lstlisting}

\begin{lstlisting}[style=textFileStyle, caption=Prompt for $\deltaf{}$]
You are given the output of an action taken by an autonomous intelligent agent navigating a web browser.  Your objective is to produce a description of the changes made to the state of the browser.

Here's the information you'll have:

Initial state of the browser as an accessibility tree: This is a simplified representation of the webpage, providing key information.
Final state of the browser: This is the accessibility tree representation after the agent took the action

The action taken by the web agent: The agent can take actions that fall under the following categories: {action_str}

To be successful, it is very important to follow the following rules:
1. Explictly think about the various features on the website and how the interaction with the website changed these features
2. Provide the description of changes in one or two sentences.
3. Strictly follow the format "State change: <your-answer>" for your output
\end{lstlisting}

\begin{lstlisting}[style=textFileStyle, caption=Prompt for the Trajectory Labeling function $\labelf{}$]
Given an instruction from a user, an autonomous intelligent agent carries out a sequence of actions on a web-browser. The actions the agent can take fall under the following categories: {action_str}

Your objective is to guess the instruction the user gave, given the following information:
Trajectory: This is a sequence of natural language descriptions of the agent's interaction with the web-browser.

Here are some examples of user instructions:
1. Get the distance from SF airport to Palo Alto.
2. Find out the price of Apple airpods
3. Add 5 items to cart
4. Make a comment on the first post in the r/gaming subreddit.

To be successful, it is very important to follow the following rules:
1. Explictly think about how the trajectory is a valid way to achieve the instruction, before outputing the instruction.
2. Start by thinking by outputing Thought: <your-reasoning>.
3. End your answer by strictly following the format "Instruction: <your-answer>" for your output.
\end{lstlisting}

\begin{lstlisting}[style=textFileStyle, caption=Prompt for the reward function $\rewardf{}$]
An autonomous intelligent agent navigating a web browser is given an instruction by a user. Your objective is to give a score to the agent based on how well it completed its task. Your score must be on the scale of 1 to 5. Give a score of 5 only when there are no errors. To do this task you are provided with the following information:

Instruction: This is the natural language instruction given to the agent.
Trajectory: This is a sequence of natural language descriptions of the agent's interaction with the web-browser.

To be successful, it is very important to follow the following rules:
1. Explictly think about what is needed to follow the instruction correctly on the website and if the trajectory reflects these steps.
2 Give a score of 4 if there are minor errors, or if the task was more than 70% completed. Give a score of 3 (or below) if the model made very little progress towards the given instruction.
3. Start by thinking by outputing Thought: <your-reasoning>.
4. End your answer by strictly following the format "Reward: <your-answer>" for your output
\end{lstlisting}

\begin{lstlisting}[style=textFileStyle, caption=Prompt for the base LLM agent $\policy$]
You are an autonomous intelligent agent tasked with navigating a web browser. You will be given web-based tasks. These tasks will be accomplished through the use of specific actions you can issue.

Here's the information you'll have:
The user's objective: This is the task you're trying to complete.
The current web page's accessibility tree: This is a simplified representation of the webpage, providing key information.
The current web page's URL: This is the page you're currently navigating.
The open tabs: These are the tabs you have open.
The previous actions: These are all the action you have performed. It may be helpful to track your progress.

The actions you can perform fall into several categories: {action_str}

To be successful, it is very important to follow the following rules:
1. You should only issue an action that is valid given the current observation
2. You should only issue one action at a time.
3. You should follow the examples to reason step by step and then issue the next action.
4. You are strictly forbidden from issuing a goto action to a URL that is not on the homepage.
5. Generate the action in the correct format. Start by reasoning about the current situation. End with "In summary, the next action I will perform is" phrase, followed by action inside ``````. For example, "Let's think step-by-step. Given the current state, I need to click on the like button which has id 1234. In summary, the next action I will perform is ```click [1234]```".
6. Issue stop action when you think you have achieved the objective. Don't generate anything after stop.

Here are some example outputs for some random tasks:
1. Let's think step-by-step. This page list the information of HP Inkjet Fax Machine, which is the product identified in the objective. Its price is $279.49. I think I have achieved the objective. I will issue the stop action with the answer. In summary, the next action I will perform is ```stop [$279.49]```
2. Let's think step-by-step. This page has a search box whose ID is [164]. According to the nominatim rule of openstreetmap, I can search for the restaurants near a location by "restaurants near". I can submit my typing by pressing the Enter afterwards. In summary, the next action I will perform is ```type [164] [restaurants near CMU] [1]```
\end{lstlisting}

Both WebArena and WebVoyager require web-agents to output a special \texttt{[stop]} action at the end of the episode. We append this stop token to \ours demonatrations via the following prompt to the base LLM.

\begin{lstlisting}[style=textFileStyle, caption=Prompt for appending the special \texttt{[stop]} action]
Given an instruction from a user, an autonomous intelligent agent carries out a sequence of actions on a web-browser. The actions the agent can take fall under the following categories (we also provide the descriptions of each action): {action_str}

You are given the user instruction, and the final webpage after the agent finished its task. Unfortunately, we forgot to collect the final stop action from the agent. Your objective is to guess the agent's stop action. To do this, you are given the following
Instruction: This is the instruction given by the user.
Final State: This is the final state of the web-page after the agent executed its actions on the browser.

Here are some examples of valid outputs:
1. Let's think step-by-step. The task requires me to find the person with the most number of upvotes. I see the answer to that is Alice Oh. Therefore I will stop now. In summary, my next action will be ```stop [Alice Oh]```.
2. Let's think step-by-step. The task required setting the price of Sprite to 25$ which I have already done. Thus I will stop now. In summary, my next action will be ```stop [N/A]```.
3. Let's think step-by-step. I was supposed to find the distance from Brad's house to the coffee shop. I see this info on the map as 0.3 miles. Thus I will issue the stop action. In summary, my next action will be ```stop [0.3 miles]```

To be successful, it is very important to follow the following rules:
1. Explictly think about what kind of a stop action was needed. For instance, if the user requests information (e.g. Search for airports near CMU or Find developers with more than 5 merge requests), then the stop action should have the answer based on the final web-page (e.g. ```stop [Pittsburgh Airport]``` or ```stop [Don Knuth, Alan Turing]```). Otherwise, the stop action should be without any arguments (e.g. ```stop```).
2. Your output should include reasoning steps. Also make sure to wrap the stop action in triple backticks for e.g. ```stop [<your answer>]```. Overall, follow the following format for your output: "Let's think step by step. <your reasoning>. In summary, my next action should be ```stop [<your answer>]```.
\end{lstlisting}

\section{Processing Demonstrations for SFT}
\label{sec:dem_postprocess_sft}

As mentioned in \S\ref{sec:background}, for supervised finetuning each demonstration is converted into multiple training instances. We perform this conversion differently based on input features of $\policy$.

\paragraph{MiniWoB++.} For MiniWoB++, $\policy$ conditions on the current observation $o_t$, the goal $g$ and the previous action $a_{t-1}$ (see prompt in \S\ref{sec:miniwob_prompts}). Thus, we pre-process each $(g, \tau)$ demonstration into inputs $(g, o_t, a_{t-1})$ with the corresponding reasoning step and action $(r_t, a_t)$ as the target output. 

\paragraph{WebArena and WebVoyager.} For WebArena and WebVoyager, $\policy$ conditions on the current observation $o_t$, the goal $g$ and all previous actions $\{a_1, a_2, \ldots, a_{t-1}\}$ (see prompt in \S\ref{sec:webarena_prompts}). Thus, we pre-process each $(g, \tau)$ demonstration into inputs $(g, o_t, \{a_{<t}\})$ with $(r_t, a_t)$ as the target output.

\section{Training Details}
\label{sec:training_details}

\paragraph{Additional Hyperparameters.} For all \llamas finetuning experiments, we set the batch size for training as $128 \times 20000$ (where 20000 is our context window), train for 2 epochs, with a learning rate of 2e-5 that linearly warms up from 0 over 3\% of total training steps. We use 4 H100 GPUs with 80GB GPU memory, and additionally use DeepSpeed ZeRO-3 \citep{rajbhandari2020zero} to speed up training and manage memory.

\section{Distribution of Intents in \ours demonstrations and Examples}
\label{sec:appendix_diversity}

\begin{figure}
    \centering
    \includegraphics[width=\linewidth]{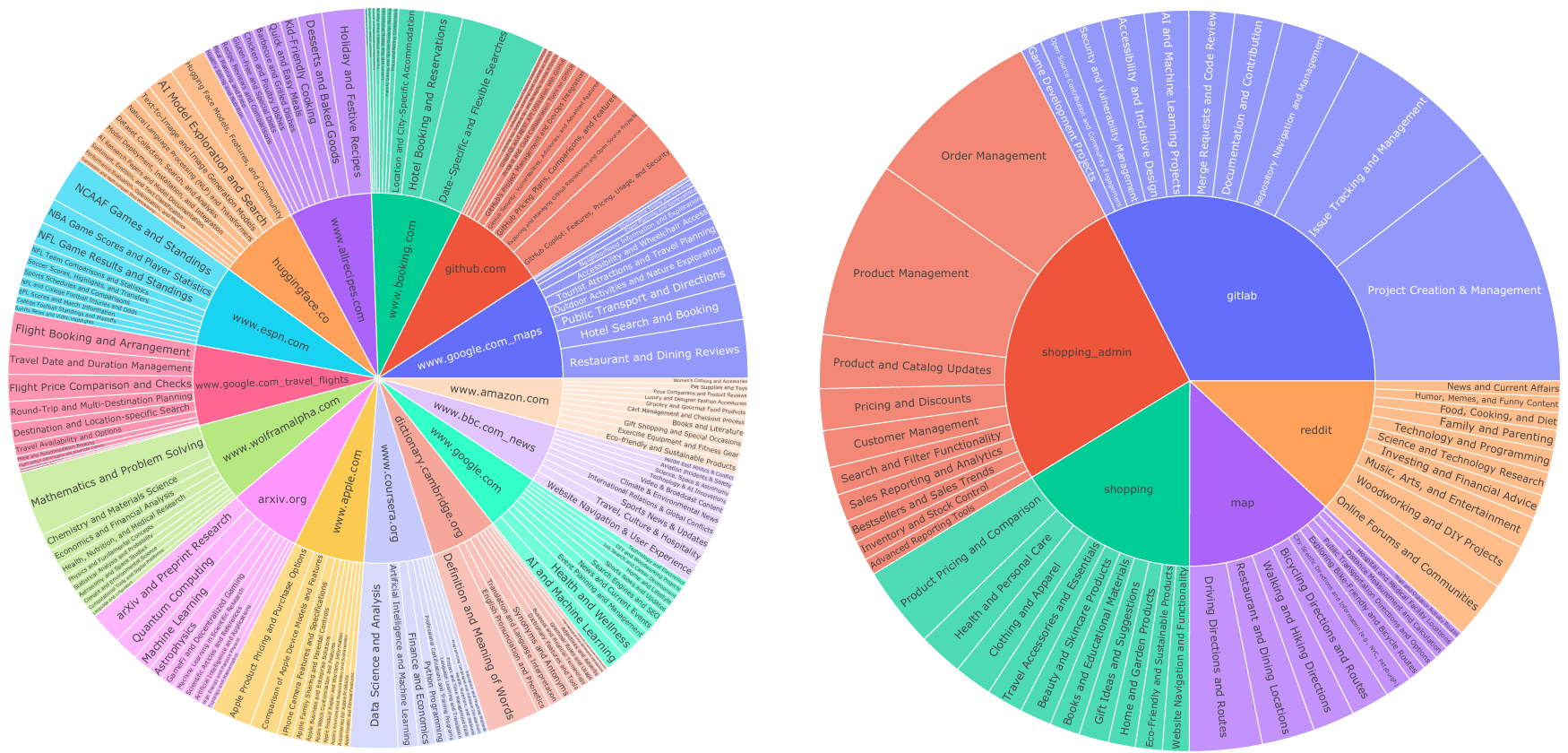}
    \caption{Top-10 intents per website for Live websites (left) and WebArena websites (right). We find a highly diverse range of intents ranging from finding \textit{holiday and festive recipes}, \textit{kid-friendly cooking}, \textit{finding restaurant and dining reviews}, \textit{finding apple product pricing} etc. Note that on live-websites, we explicitly prevent models from logging in, and this inherently limits the kinds of tasks it can do. No such limitations are placed on WebArena, leading to tasks that require logging in such as itextit{posting on forums}, \textit{creating projects}, \textit{managing order details} etc. We report the perplexity of intent distribution per website in Section~\ref{sec:nnetnav_details}}
    \label{fig:intent_burst_plot}
\end{figure}

\begin{table}[t]
\centering
\small
\renewcommand{\arraystretch}{1.2}
\begin{tabularx}{\textwidth}{|X|}
\toprule
\hrow \multicolumn{1}{|c|}{\it Shopping} \\ \midrule
Find a kitchen utensil organizer. \\
Find a kitchen utensil organizer within a certain budget.  \\
Write a review for the product ``Citric Acid 2 Pounds 100\% Pure Organic Food Grade''. \\
Find the price of kitchen gadgets that can be used for dining and entertaining, and add them to the cart. \\
Browse for women's clothing items, specifically jumpsuits, and add some to cart. \\ 
\hrow \multicolumn{1}{|c|}{\it CMS} \\ \midrule
Change the stock status of the Sprite Stasis Ball 65 cm to In Stock. \\
Create a new product in the Magento Admin panel with the name 'New Fashionable Watch', SKU 'New Fashionable WatchFW101', price \$100.00, and set as new from 2024-01-01. \\
Update the price of Sprite Stasis Ball 55 cm to \$24.50 and set its quantity to 50.  \\
Add two products, ``Abominable Hoodie'' and ``Samsung Smart TV'', with respective prices \$99.99 and \$50.00, and then start the process of adding a new customer. \\
\hrow \multicolumn{1}{|c|}{\it Reddit} \\ \midrule
Create a new forum called ``Funny Stuff'' with the title ``Memes and LOLs'', description ``A place for sharing and discussing funny memes and LOLs'', and sidebar ``Memes of the day''. \\
Find a webpage related to intraday trading strategies on the wallstreetbets forum. \\
Find and participate in a discussion on the wallstreetbets forum about intraday trading strategy, specifically on a post titled ``Swings and roundabouts''. \\ 
Change my profile settings to use Deutsch as the language and Africa/Accra as the time zone, and then view the search results for ``r/art''. \\ 
\hrow \multicolumn{1}{|c|}{\it Maps} \\ \midrule
Get walking directions from Logan Street, Pittsburgh, PA to Carnegie Mellon University on OpenStreetMap. \\
Get the cycling directions from Brooklyn to Manhattan. \\
Find the driving directions from TLC Medical Transportation Services in Syracuse to Times Square in Manhattan. \\
\hrow \multicolumn{1}{|c|}{\it Gitlab} \\ \midrule
Create a new project named 'My Blog Post Project' and add an Apache License 2.0 file.\\
Create a new project, add a LICENSE file with Apache License 2.0, and approve the ``Add verification functions'' merge request.\\ 
Search for a README.md file within the ``My New Project'' project and verify its contents. \\
Edit the issue ``Link to WCAG 2.1 instead of 2.0?'' in the First Contributions project on GitLab by updating its title and description to point to WCAG 2.1 guidelines instead of 2.0 guidelines.\\
Investigate the node-http-proxy project's issue \#992 regarding connection headers and determine its relevance to the Byte Blaze project. \\
Investigate and comment on the ``Outdated dependencies'' issue in the ``Byte BlazeByte BlazeByte Blaze / accessible-html-content-patterns'' project. \\

\bottomrule
\end{tabularx}
\caption{Some Example demonstrations obtained from \ours-WA. We note that these instructions are hierarchical, refer to concrete features and entities and plausible by design.}
\label{tab:qual_examples}
\end{table}

\end{document}